\documentclass[lettersize,journal]{IEEEtran}
\usepackage{amsmath,amsfonts}
\usepackage{array}
\usepackage[caption=false,font=normalsize,labelfont=sf,textfont=sf]{subfig}
\usepackage{textcomp}
\usepackage{stfloats}
\usepackage{url}
\usepackage{verbatim}
\usepackage{graphicx}
\usepackage{cite}
\usepackage{pifont}
\usepackage{authblk}

\usepackage{amsthm}

\usepackage{booktabs}
\usepackage{multirow}
\usepackage{lipsum}
\usepackage{hyperref}
\usepackage{breakurl}

\usepackage[edges]{forest}
\definecolor{hidden-draw}{RGB}{20,68,106}
\definecolor{hidden-pink}{RGB}{255,245,247}
\usepackage[framemethod=tikz]{mdframed}
\usepackage{subcaption}
\usepackage{soul}
\usepackage{times}
\usepackage{tabularx}
\usepackage{float}
\usepackage{xcolor} 
\usepackage{latexsym}
\usepackage{bbm}
\usepackage{amsfonts}
\usepackage{dsfont}
\usepackage{graphicx}
\usepackage{caption}
\usepackage{subcaption}
\usepackage{algpseudocode}
\usepackage{colortbl}
\definecolor{lightgray}{gray}{0.9}
\definecolor{lightgreen}{rgb}{0.9, 1, 0.9}
\usepackage{microtype}
\usepackage{inconsolata}

\usepackage{amsmath,amsfonts,bm}

\def\eqref#1{equation~\ref{#1}}

\def\1{\bm{1}}

\DeclareMathAlphabet{\mathsfit}{\encodingdefault}{\sfdefault}{m}{sl}
\SetMathAlphabet{\mathsfit}{bold}{\encodingdefault}{\sfdefault}{bx}{n}

\usepackage{tikz}
\usetikzlibrary{fadings}
\usetikzlibrary{mindmap,trees}
\usetikzlibrary{arrows,automata,shapes,positioning,shadows,trees}
\usetikzlibrary{shapes,snakes,shadows}
\usetikzlibrary{shapes.arrows}
\usetikzlibrary{calc,shapes, positioning}
\usetikzlibrary{decorations.text}
\usetikzlibrary{matrix,chains,positioning,decorations.pathreplacing,arrows}
\usetikzlibrary{bayesnet}
\usetikzlibrary{shadows.blur}
\usetikzlibrary{shapes.geometric}

\usepackage{pgfplots}
\usepackage{pgfplotstable}
\usepackage{pgfmath,pgffor}
\usepgfplotslibrary{colorbrewer}
\pgfplotsset{compat = 1.14, cycle list/Set1-8}
\usetikzlibrary{pgfplots.statistics, pgfplots.colorbrewer}
\pgfplotsset{compat=1.8}

\tikzstyle{edge}=[-latex',draw=black!90,shorten <=1pt,shorten >=1pt]
\tikzstyle{redge}=[latex'-,draw=black!90,shorten <=1pt,shorten >=1pt]
\tikzstyle{dedge}=[latex'-latex',draw=black!90,shorten <=1pt,shorten >=1pt]

\tikzstyle{block}=[draw, text width=5em,align=center,shape=rectangle, rounded corners, , align=center]
\tikzstyle{nobox}=[align=center]
\definecolor{emb}{RGB}{209,228,252}
\definecolor{hidden-blue}{RGB}{194,232,247}
\definecolor{hidden-orange}{RGB}{224,224,224}
\definecolor{hidden-yellow}{RGB}{242,244,193}
\definecolor{output-purple}{RGB}{219,203,231}
\definecolor{output-green}{RGB}{204,231,207}
\definecolor{hiddendraw}{RGB}{10,128,122}
\definecolor{myred2}{HTML}{F875AA}
\definecolor{mypurple2}{HTML}{D2E0FB}

\definecolor{myred}{HTML}{F8F6F4}
\definecolor{mypurple}{HTML}{FFDFDF}
\definecolor{myyellow}{HTML}{FFF6F6}
\definecolor{mygreen}{HTML}{D2E0FB}
\usepackage{todonotes}

\usepackage{todonotes}

\newcommand{\cmark}{\ding{51}}%
\newcommand{\xmark}{\ding{55}}%

\begin{document}

\title{A Survey of Graph Retrieval-Augmented Generation for Customized Large Language Models}

\author{
\IEEEauthorblockN{
\textbf{
Qinggang~Zhang$^*$, Shengyuan~Chen$^*$, Yuanchen~Bei$^*$, Zheng~Yuan, Huachi~Zhou
}}

\IEEEauthorblockN{\textbf{Zijin~Hong, Hao~Chen, Yilin Xiao, Chuang Zhou, Junnan~Dong, Yi~Chang, Xiao~Huang}}

\thanks{*Authors contributed equally to this research.}
\thanks{Qinggang Zhang, Shengyuan Chen, Yuanchen Bei, Zheng Yuan, Huachi Zhou, Zijin Hong, Hao Chen, Yilin Xiao, Chuang Zhou, Junnan Dong, and Xiao Huang are with the The Hong Kong Polytechnic University, Hong Kong SAR, China (e-mail: \{qinggangg.zhang, yzheng.yuan, huachi.zhou, zijin.hong, yilin.xiao, chuang-qqzj.zhou, hanson.dong\}@connect.polyu.hk, \{iyuanchenbei, sundaychenhao\}@gmail.com, \{sheng-yuan.chen, xiao.huang\}@polyu.edu.hk).}
\thanks{Yi Chang is with the School of Artificial Intelligence, Jilin University, Changchun, China (e-mail: yichang@jlu.edu.cn).}
}

\maketitle

\begin{abstract}

Large language models (LLMs) have demonstrated remarkable capabilities in a wide range of tasks, yet their application to specialized domains remains challenging due to the need for deep expertise. Retrieval-augmented generation (RAG) has emerged as a promising solution to customize LLMs for professional fields by seamlessly integrating external knowledge bases, enabling real-time access to domain-specific expertise during inference. Despite its potential, traditional RAG systems, based on flat text retrieval, face three critical challenges: (i) complex query understanding in professional contexts, (ii) difficulties in knowledge integration across distributed sources, and (iii) system efficiency bottlenecks at scale. This survey presents a systematic analysis of Graph-based Retrieval-Augmented Generation (GraphRAG), a new paradigm that revolutionizes domain-specific LLM applications. GraphRAG addresses traditional RAG limitations through three key innovations: (i)  graph-structured knowledge representation that explicitly captures entity relationships and domain hierarchies, (ii) efficient graph-based retrieval techniques that enable context-preserving knowledge retrieval with multihop reasoning ability, and (iii) structure-aware knowledge integration algorithms that leverage retrieved knowledge for accurate and logical coherent generation of LLMs.
In this survey, we systematically analyze the technical foundations of GraphRAG and examine current implementations across various professional domains, identifying key technical challenges and promising research directions. All the related resources of GraphRAG, including research papers, open-source data, and projects, are collected for the community in \textcolor{blue}{\url{https://github.com/DEEP-PolyU/Awesome-GraphRAG}}.

\end{abstract}


\ifCLASSOPTIONcompsoc
\IEEEraisesectionheading{\section{Introduction}\label{sec:introduction}}
\else

\section{Introduction}
\label{sec:introduction}
\begin{figure*}[tbp]
\vspace{-5mm}
    \centering
    \includegraphics[width=1.\linewidth, trim=0cm 0cm 0cm 0cm,clip]{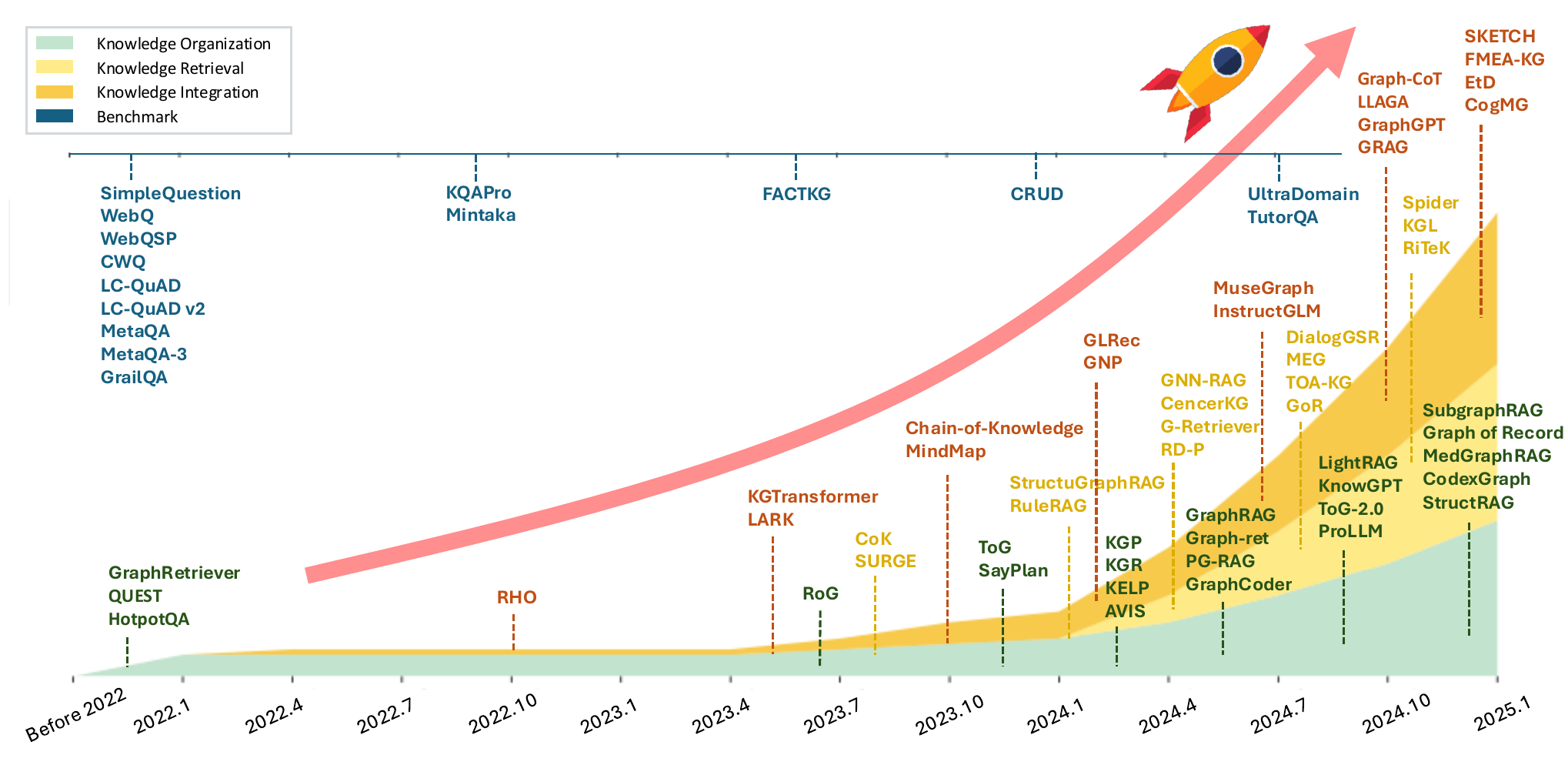}
    \vspace{-8mm}
    \caption{The development trends in the field of GraphRAG with representative works.}
    \vspace{-5mm}
    \label{fig:stream_graph}
\end{figure*}

Large language models (LLMs), like the GPT series~\cite{openai2023gpt4}, have surprised the world with their remarkable ability across a wide range of tasks, achieving breakthroughs in text comprehension~\cite{brown2020language}, question answering~\cite{khashabi2020unifiedqa}, and content generation~\cite{chowdhery2023palm}. Despite their effectiveness, LLMs are always criticized for their limited ability to handle knowledge-intensive tasks, especially when faced with questions requiring domain expertise~\cite{Zhang-etal24KnowGPT}.  Specifically, applying LLMs into specialized domains remains challenging for three fundamental reasons, including \ding{182} Knowledge limitations: LLMs' pre-trained knowledge is broad but shallow in specialized fields. Their training data primarily consists of general-domain content, leading to insufficient depth in professional domains and potential inconsistencies with current domain-specific standards and practices. \ding{183} Reasoning complexity: Specialized domains require precise, multi-step reasoning with domain-specific rules and constraints. LLMs often struggle to maintain logical consistency and professional accuracy throughout extended reasoning chains, particularly when dealing with technical constraints or domain-specific protocols. \ding{184} Context sensitivity: Professional fields often involve context-dependent interpretations where the same terms or concepts may have different meanings or implications based on specific circumstances. LLMs frequently fail to capture these nuanced contextual variations, leading to potential misinterpretations or inappropriate generalizations.

To adapt LLMs for specific or private domains, initial strategies involved {\bf fine-tuning LLMs} with specialized datasets~\cite{hu2022lora}. This method enhances performance by adding a limited number of parameters while fixing the parameters learned in the pre-training~\cite{he2022towards}. However, the significant distribution gap between the domain-specific dataset and the pre-training corpus makes it challenging for LLMs to integrate new knowledge without compromising their existing understanding~\cite{gekhman2024does}. A recent study by Google Research further highlighted the risks associated with using supervised fine-tuning to update knowledge, particularly in cases where new knowledge conflicts with pre-existing information; acquiring new knowledge through supervised fine-tuning can lead to the model generating new hallucinations and even experiencing severe catastrophic forgetting~\cite{zhai2024investigating}.

\textbf{Retrieval-augmented generation (RAG)} offers a promising solution to customize LLMs for specific domains~\cite{lewis2020retrieval,zhang2025faithfulrag}. Rather than retraining LLMs to incorporate updates, RAG enhances these models by leveraging external knowledge from text corpora without modifying their architecture or parameters.  This approach enables LLMs to generate responses by leveraging not only their pre-trained knowledge but also real-time retrieved domain-specific information, thereby providing more accurate and reliable answers. The naive RAG systems operate through three key steps: knowledge preparation, retrieval, and integration. During knowledge preparation, the external textual corpus is divided into manageable textual chunks and converted into vector representations for efficient indexing. In the retrieval stage, when a user submits a query, the system searches for relevant chunks using keyword matching or vector similarity measures. The integration stage then combines these retrieved chunks with the original query to generate more informed responses. Recently, some advanced RAG systems have evolved beyond simple text chunk retrieval to offer more sophisticated knowledge augmentation approaches. These include hierarchical RAG that preserves document structure through multi-level retrieval~\cite{chen2024multi,li2024structrag}, re-ranking systems that implement two-stage retrieval for higher recall and precision~\cite{glass2022re2g,xu2023recomp}, self-querying RAG that automatically decomposes complex queries~\cite{asai2023self}, and adaptive RAG that dynamically adjusts retrieval strategies based on query types~\cite{tang2024mba,sarthi2024raptor}. These strategies advance naive RAG systems by improving context awareness, retrieval accuracy, and handling complex queries.

The emergence of RAG has offered a promising approach for customizing LLMs with domain-specific knowledge. However, despite its potential, RAG faces several critical limitations that impact its effectiveness in practical applications. These limitations can be broadly categorized into four main challenges that significantly affect the performance and utility of RAG-enhanced LLMs.
The primary challenge lies in \textbf{complex query understanding}. Specialized domains often involve intricate terminology and industry-specific jargon that requires precise interpretation~\cite{ling2023domain}. User queries in these areas typically contain numerous technical terms and industry-specific expressions, with solutions often requiring reasoning across multiple related concepts. 
Traditional RAG approaches, which rely on simple keyword matching and vector similarity techniques, are inadequate for capturing the deep semantic nuances and multi-step reasoning processes necessary for accurate and comprehensive~\cite{bruckhaus2024rag}. For instance, when queried about the connection between concept A and concept D, these systems typically retrieve only directly related information, missing crucial intermediate concepts like B and C that could bridge the relationship. This narrow retrieval scope limits RAG's ability to a broad contextual understanding and complex reasoning.

Another key challenge involves \textbf{integrating domain knowledge from distributed sources}.
Domain knowledge is usually collected from different sources, such as textbooks, research papers, industry reports, technical manuals, and maintenance logs. These textual documents may have varying levels of quality, accuracy, and completeness. The retrieved knowledge is often flattened, extensive, and intricate, while domain concepts are typically scattered across multiple documents without clear hierarchical relationships between different concepts~\cite{Zhang-etal24KnowGPT,sun2024thinkongraph,ma2024think-on-graph-2.0}. Although RAG systems attempt to manage this complexity by dividing documents into smaller chunks for effective and efficient indexing, this approach inadvertently sacrifices crucial contextual information, significantly compromising retrieval accuracy and contextual comprehension. 
This limitation hampers the ability to establish robust connections between related knowledge points, leading to fragmented understanding and reduced efficacy in leveraging domain-specific expertise.
 
The third limitation stems from the \textbf{inherent constraints of LLMs}. While RAG systems can retrieve relevant information from vast knowledge bases, the LLM's ability to process this information is constrained by its fixed context window (typically 2K-32K tokens)~\cite{openai2023gpt4,anthropic2024claude}. Long-range dependencies in complex documents cannot be fully captured, as the content exceeding the context window must be truncated or summarized, disrupting natural semantic units and logical flow. The challenge of maintaining coherence across extensive knowledge contexts becomes increasingly problematic in professional domains, as critical information may be lost during context window truncation. This fundamental limitation directly impacts the system's ability to process and synthesize comprehensive information from large-scale knowledge bases.

The last challenge relates to system \textbf{efficiency and scalability}. The entire RAG pipeline - from initial corpus preprocessing and indexing to real-time retrieval and generation - faces significant efficiency bottlenecks~\cite{edge2024local,guo2024lightrag}. The external knowledge base contains a lot of domain-irrelevant information, while domain-specific terminologies are always sparsely distributed over these documents. RAG systems can be computationally expensive and time-consuming~\cite{edge2024local}, especially when dealing with large-scale knowledge sources, as the model needs to search through vast amounts of unstructured text to find relevant information. 
Moreover, real-time retrieval and cross-document reasoning can introduce considerable latency, negatively impacting user experience. The scalability of RAG is further constrained by declining retrieval quality and accuracy as the size of the knowledge base grows~\cite{guo2024lightrag}, thereby limiting its practical deployment in extensive and dynamic environments.

To address these limitations, \textbf{graph retrieval-augmented generation (GraphRAG)} has recently emerged as a new paradigm to customize LLMs with well-organized background knowledge and improved contextual reasoning~\cite{edge2024local,liang2024kag,peng2024graph,jeon2025hybrid,procko2024graph}. 
Based on the utilization of graph structures, existing GraphRAG models can be categorized into three main categories: \ding{182} \textbf{Knowledge-based GraphRAG}, which uses graphs as knowledge carriers. \ding{183} \textbf{Index-based GraphRAG}  uses graphs as index tools to retrieve relevant raw texts from the corpus, and \ding{184} \textbf{Hybrid GraphRAG} which combines the strengths of both knowledge-based and index-based frameworks, providing more advanced solutions for complex reasoning tasks. 
Knowledge-based and Index-based GraphRAG represent two distinct paradigms for enhancing LLMs with structured knowledge. Knowledge-based GraphRAG focuses on transforming unstructured textual documents into explicit and structured KGs, where nodes represent domain concepts and edges capture semantic relationships between them, enabling better representation of hierarchical relationships and complex knowledge dependencies. In contrast, Index-based GraphRAG maintains the original textual form while utilizing graph structures primarily as an indexing mechanism to organize and retrieve relevant text chunks efficiently. By incorporating graph structures into text indexing, Index-based GraphRAG methods establish semantic connections between text chunks for efficient look-up operations and retrieval. While Knowledge-based GraphRAG emphasizes the explicit modeling of domain knowledge and semantic relationships through graph transformation, Index-based GraphRAG prioritizes efficient information retrieval and global navigation through the graph-based organization of the raw text. This fundamental difference in approach reflects their distinct purposes: Knowledge-based GraphRAG aims to create a structured knowledge representation for a better understanding of complex relationships with graph-based reasoning capability, whereas Index-based GraphRAG focuses on optimizing the retrieval and accessibility of relevant textual information through graph-based indexing strategies. 

In this survey, we systematically analyze the technical foundations of GraphRAG and examine current implementations across various professional domains, identifying key technical challenges and promising research directions. All the related resources of GraphRAG, including research papers, open-source data, and projects, are collected for the community in \textcolor{blue}{\url{https://github.com/DEEP-PolyU/Awesome-GraphRAG}}.

\begin{figure*}[tbp]
    \centering
    \includegraphics[width=1.\linewidth, trim=0cm 0cm 0cm 0cm,clip]{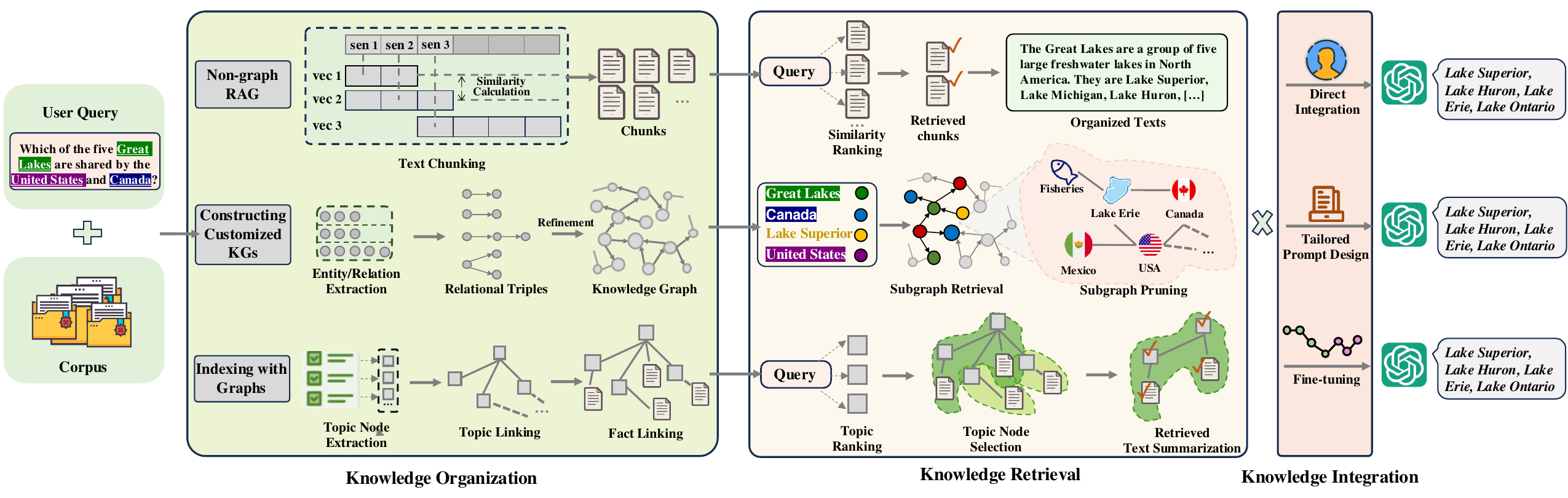}
    \vspace{-5mm}
    \caption{A comprehensive overview of traditional RAG and two typical GraphRAG workflows. Non-graph RAG organizes the corpus into chunks, ranks them by similarity, and retrieves the most relevant text for generating responses. Knowledge-based GraphRAG extracts detailed knowledge graphs from the corpus using entity recognition and relation extraction, offering fine-grained, domain-specific information. Index-based GraphRAG summarizes the corpus into high-level topic nodes, which are linked to form an index graph while fact-linking maps topics to text. This two-layer structure combines efficient topic retrieval with detailed text knowledge, offering scalability and performance compared to Knowledge-based GraphRAG. }
    \vspace{-5mm}
    \label{fig:taxonomy_fig}
\end{figure*}

\section{Overview}\label{sec:overview}
This survey provides a comprehensive analysis of GraphRAG, detailing its taxonomy, mechanisms, challenges, and future research directions, which is organized into seven main sections that progress from foundational concepts to practical implementations. Specifically, we begin in Section~\ref{sec:def} by establishing the foundational framework, tracing GraphRAG's evolution from traditional RAG systems, examining RAG's limitations in handling structured knowledge, and introducing GraphRAG's core concepts and advantages in complex reasoning tasks. The following three sections systematically explore the key components of GraphRAG systems: two primary paradigms for structured knowledge organization, including knowledge carrier graphs and index graphs (Section~\ref{sec:organization}), retrieval techniques for extracting query-relevant factual information from structured knowledge bases (Section~\ref{sec:retrieval}), and knowledge integration methods for effectively incorporating retrieved knowledge into LLMs (Section~\ref{sec:integration}). Moving toward practical applications, Section~\ref{sec:opensource} (see Appendix) explores the real-world implementations of GraphRAG systems by reviewing open-source projects and benchmark datasets across different domains. Finally, Sections~\ref{sec:future} and~\ref{sec:discussion} (see Appendix) conclude by identifying future research directions and discussing potential challenges in knowledge quality, retrieval efficiency, system generalization, and security. Throughout the survey, we balance fundamental concepts, the current state-of-the-art methods, and practical implementations, making it valuable for both researchers advancing the field and practitioners developing GraphRAG applications in real-world scenarios. 

Our survey advances beyond the existing survey~\cite{peng2024graph,procko2024graph,han2024retrieval} through a more systematic and comprehensive review of existing GraphRAG systems. While the previous survey offered a basic workflow description covering Graph-based Indexing, Graph-guided Retrieval, and Graph-enhanced Generation, we introduce a more sophisticated and comprehensive taxonomy that clearly categorizes GraphRAG approaches into three distinct categories (Knowledge-based, Index-based, and Hybrid GraphRAG), providing a more nuanced understanding of the field. Our survey features a more systematic six-section structure that progresses logically from theoretical foundations to practical implementations, offering a more detailed exploration of each component, including knowledge organization paradigms, retrieval techniques, and integration methods. Unlike the previous survey, we provide extensive practical guidance through a detailed review of open-source projects, and domain-specific case studies supported by comprehensive datasets across different domains. We also offer a more thorough analysis of challenges and solutions across multiple dimensions, including knowledge quality, retrieval efficiency, system generalization, and security concerns. Finally, while the existing survey broadly discussed potential applications, we provide actionable insights supported by comprehensive analysis and implementation examples, making our survey a more valuable resource for practitioners to deploy GraphRAG systems in production environments.

\section{What is GraphRAG}\label{sec:def}
This section provides an overview of RAG with LLMs, including the traditional RAG pipeline, the GraphRAG, and the advantages GraphRAG offers over traditional RAG systems.

\subsection{Traditional RAG Pipeline}
A RAG framework begins by retrieving relevant information from pre-constructed external knowledge bases based on the query. This information is then used to prompt LLMs, guiding them in constructing credible reasoning chains. As a result, RAG enables LLMs to generate more substantiated and accurate content, effectively minimizing hallucinations and inconsistencies. The traditional RAG pipeline typically comprises three core components: knowledge organization, knowledge retrieval, and knowledge integration. We provide a detailed analysis of traditional RAG systems in Appendix D.

\subsubsection{Knowledge organization}
In traditional RAG systems, knowledge organization involves structuring and preparing external knowledge repositories to facilitate rapid and relevant retrieval when provided with a query. A common strategy is to split the large-scale text corpus into manageable chunks. These chunks are then transformed into embeddings using an embedding model, where the embeddings serve as keys of original text chunks in a vector database~\cite{borgeaud2022improving,izacard2023atlas,jiang2023active}. This setup enables efficient look-up operations and retrieval of relevant content via distance-based search in the semantic space.

As a crucial step in the pre-retrieval process, several methods~\cite{gao2023retrieval,wang2024searchingforbestpractivesinrag} have been proposed to optimize knowledge organization, focusing on two main aspects: granularity optimization and indexing optimization. Granularity optimization aims to balance relevance and efficiency, as coarse-grained units provide richer context but risk redundancy and distraction, while fine-grained units may lack semantic integrity and increase the retrieval burden~\cite{yu2023chain,zhong2024mixofgranularity}. To control granularity, chunking strategies are employed to split documents into chunks based on token limits. Methods such as recursive splits, sliding windows, and Small-to-Big~\cite{wang2024searchingforbestpractivesinrag,llamaindex_website} strive to maintain semantic completeness while optimizing context length. Indexing optimization seeks to improve the structure and quality of content for retrieval. Metadata-addition techniques, which attach chunk text with metadata like titles, timestamps, categories, and keywords, enable filtering and re-ranking operations during the post-retrieval process~\cite{wang2024searchingforbestpractivesinrag}. Another type of technique is hierarchical indexing, which organizes files into parent-child relationships with summaries at each node, facilitating faster and more efficient data traversal while reducing retrieval errors~\cite{wang2024corag}. Such tree-like indexing methods represent early attempts at structured knowledge organization and have inspired successors to harness the power of graph structures for knowledge organization, i.e., GraphRAG.

In summary, knowledge organization is foundational to the retrieval process. By carefully constructing the knowledge resources, RAG systems can improve the retrieval efficiency and ensure the fidelity and reliability of retrieved content.

\subsubsection{Knowledge retrieval}
The knowledge retrieval stage encompasses various methods and strategies designed to efficiently access and retrieve the necessary knowledge from pre-organized repositories, ensuring the selection of relevant information that can enhance the quality of generated outputs.

Current RAG systems usually involve retrieval methods such as k-nearest neighbor retrieval (KNN), term frequency-inverse document frequency (TF-IDF), and best matching 25 (BM25) to retrieve the relevant content.
RETRO~\cite{borgeaud2022improving} employs KNN to extract approximate relevant neighbors from the conducted key-value database by calculating the L2 distance. RETROprompt~\cite{chen2022decoupling} extends this approach with a few-shot knowledge store, tailoring it for more advanced prompts.

In addition, some specialized techniques are used prior to the retrieval method to improve the accuracy and efficiency of the retrieval. To capture multifaceted aspects of the query, GAR~\cite{mao2020generation} introduces diverse context generation, enriching the initial query with additional contexts before applying BM25 retrieval. Enhancing this framework, EAR~\cite{chuang2023expand} implements a re-ranking process that selects the optimal candidate from multiple expanded queries to improve the retrieval accuracy. Furthermore, in tackling the computational challenges associated with exact retrieval methods like BM25, Doostmohammadi et al.~\cite{doostmohammadi2023surface} propose a hybrid approach to identify approximate neighbors using sentence transformers for representation and then apply BM25 for re-ranking, effectively balancing accuracy with computational efficiency on large-scale retrieval tasks.



Overall, knowledge retrieval combines innovative methodologies with diverse data-sourcing strategies, underpinning the generation of informed and contextually relevant outputs.

\tikzstyle{my-box}=[
    rectangle,
    draw=hidden-draw,
    rounded corners,
    align=left,
    text opacity=1,
    minimum height=1.5em,
    minimum width=5em,
    inner sep=2pt,
    fill opacity=.8,
    line width=0.8pt,
]
\tikzstyle{leaf-head}=[my-box, minimum height=1.5em,
    draw=gray!80, 
    fill=gray!15,  
    text=black, font=\normalsize,
    inner xsep=2pt,
    inner ysep=4pt,
    line width=0.8pt,
]

\tikzstyle{leaf-datasets}=[my-box, minimum height=1.5em,
    draw=red!70, 
    fill=red!15,  
    text=black, font=\normalsize,
    inner xsep=2pt,
    inner ysep=4pt,
    line width=0.8pt,
]

\tikzstyle{leaf-methods}=[my-box, minimum height=1.5em,
    draw=cyan!70, 
    fill=cyan!15,  
    text=black, font=\normalsize,
    inner xsep=2pt,
    inner ysep=4pt,
    line width=0.8pt,
]
\tikzstyle{leaf-metrics}=[my-box, minimum height=1.5em,
    draw=orange!80, 
    fill=orange!15,  
    text=black, font=\normalsize,
    inner xsep=2pt,
    inner ysep=4pt,
    line width=0.8pt,
]

\tikzstyle{modelnode-datasets}=[my-box, minimum height=1.5em,
    draw=red!80, 
    fill=white,  
    text=black, font=\normalsize,
    inner xsep=2pt,
    inner ysep=4pt,
    line width=0.8pt,
]

\tikzstyle{modelnode-methods}=[my-box, minimum height=1.5em,
    draw=cyan!100, 
    fill=white,  
    text=black, font=\normalsize,
    inner xsep=2pt,
    inner ysep=4pt,
    line width=0.8pt,
]
\tikzstyle{modelnode-metrics}=[my-box, minimum height=1.5em,
    draw=orange!100, 
    fill=white,  
    text=black, font=\normalsize,
    inner xsep=2pt,
    inner ysep=4pt,
    line width=0.8pt,
]

\begin{figure*}[!th]
    \centering
    \vspace{-5mm}
    \resizebox{1\textwidth}{!}{
        \begin{forest}
            forked edges,
            for tree={
                grow=east,
                reversed=true,
                anchor=base west,
                parent anchor=east,
                child anchor=west,
                base=left,
                font=\normalsize,
                rectangle,
                draw=hidden-draw,
                rounded corners,
                align=left,
                minimum width=1em,
                edge+={darkgray, line width=1pt},
                s sep=3pt,
                inner xsep=0pt,
                inner ysep=3pt,
                line width=0.8pt,
                ver/.style={rotate=90, child anchor=north, parent anchor=south, anchor=center},
            }, 
            [
                GraphRAG for Customized LLMs, leaf-head, ver
                [
                    Knowledge \\ Organization  \\(\S\ref{sec:organization}), leaf-datasets, text width=6em
                    [ 
                        Graphs for\\Knowledge \\Indexing, leaf-datasets, text width=6em
                        [GNN-ret~\cite{li2024graph}{,} PG-RAG~\cite{liang2024empowering}{,} KGP~\cite{wang2024knowledge}{,} GraphCoder~\cite{liu2024graphcoder}{,} AVIS\cite{hu2024avis}{,} SayPlan~\cite{rana2023sayplan} HotpotQA~\cite{xiong2020mdr}{,} GraphRetriever~\cite{min2019knowledge}{,}
                        \\ KAR~\cite{liang2024kag}{,} GRAG~\cite{hu2024grag}{,}  KET-RAG~\cite{huang2025ket}{,} PIKE-RAG~\cite{wang2025pike}{,} OG-RAG~\cite{sharma2024og}{,} ArchRAG~\cite{wang2025archrag}{,} RAPTOR~\cite{sarthi2024raptor}{,} HippoRAG2~\cite{gutierrezrag}, text width=50.7em,modelnode-datasets]
                    ]
                    [
                        Graphs as \\ Knowledge\\Carriers, leaf-datasets, text width=6em
                        [
                            Knowledge Graph \\ Construction from Corpus, leaf-datasets, text width=11em
                            [StructRAG\cite{li2024structrag}{,}
                            GraphRAG\cite{edge2024local}{,}
                            FastRAG~\cite{abane2024fastrag}{,}
                            QUEST\cite{quest-quasi}{,}  
                            Structure-guided Prompt\cite{cheng2024structure}
                            \\LightRAG\cite{guo2024lightrag}{,}
                             MedRAG~\cite{zhao2025medrag}{,} GraphReader~\cite{li2024graphreader}{,} LuminiRAG~\cite{martis2024luminirag}{,} HippoRAG~\cite{jimenez2024hipporag}, modelnode-datasets, text width=38em]
                        ]
                        [
                            GraphRAG with Existing\\Knowledge Graphs, leaf-datasets, text width=11em
                            [KnowGPT~\cite{Zhang-etal24KnowGPT}{,}
                            ToG~\cite{sun2024thinkongraph}{,}
                            ToG-2.0~\cite{ma2024think-on-graph-2.0}{,}
                            RoG~\cite{luo2024reasoning}{,} KGR~\cite{guan2024mitigating}{,}  KELP~\cite{liu2024knowledge}{,} ProLLM~\cite{jin2024prollm}{,} \\ SubGraphRAG~\cite{li2024simple}{,} PoG~\cite{chen2024plan}{,} LEGO-GraphRAG~\cite{cao2024lego}{,} KnowNet~\cite{yan2024knownet}, modelnode-datasets, text width=38em]
                        ]
                    ]
                    [ 
                        Hybrid \\ GraphRAG, leaf-datasets, text width=6em
                        [GoR~\cite{zhang2024graphofrecords}{,} MedGraphRAG~\cite{wu2024medical}{,} KG2RAG~\cite{zhu2025knowledge}{,} HYBGRAG~\cite{lee2024hybgrag}{,} CodexGraph~\cite{liu2024codexgraph}, modelnode-datasets, text width=50.7em]
                    ]
                ]
                [
                    Knowledge  \\ Retrieval \\ (\S\ref{sec:retrieval}), leaf-metrics,text width=6em
                    [ 
                        Retrieval \\  Technique, leaf-metrics, text width=6em
                        [
                            Similarity-based Retriever, leaf-metrics, text width=11em
                            [PG-RAG~\cite{liang2024empowering}{,}
                            GraphCoder~\cite{liu2024graphcoder}{,}
                            MedGraphRAG~\cite{wu2024medical}{,}
                            StructuGraphRAG~\cite{zhu2024structugraphrag}{,}
                            \\CancerKG~\cite{gubanov2024cancerkg}{,}
                            G-Retriever~\cite{he2024g}, modelnode-metrics, text width=38em]
                        ]
                        [
                            Logical-based Retriever, leaf-metrics, text width=11em
                            [RoG~\cite{luo2024reasoning}{,}
                            RD-P~\cite{huang2024rd}{,}
                            RuleRAG~\cite{chen2024rulerag}{,}
                            KGL~\cite{yang2024intelligent}{,}
                            RiTeK~\cite{huang2024ritek}, modelnode-metrics, text width=38em]
                        ]
                        [
                            GNN-based Retriever, leaf-metrics, text width=11em
                            [GNN-Ret~\cite{li2024graph}{,}
                            SURGE~\cite{kang2023knowledge}{,}
                            GNN-RAG~\cite{dong2024advanced}, modelnode-metrics, text width=38em]
                        ]
                        [
                            LLM-based Retriever, leaf-metrics, text width=11em
                            [ToG~\cite{sun2024thinkongraph}{,}
                            LightRAG~\cite{guo2024lightrag}{,}
                            KGP~\cite{wang2024knowledge}{,}
                            MEG~\cite{cabello2024meg}{,}
                            TQA-KG~\cite{he2024enhancing}{,} E$^2$GraphRAG~\cite{zhao20252graphrag}, modelnode-metrics, text width=38em]
                        ]
                        [
                            RL-based Retriever, leaf-metrics, text width=11em
                            [KnowGPT~\cite{Zhang-etal24KnowGPT}{,}
                            Spider~\cite{shen2025insight}{,} GraphRAG-R1~\cite{yu2025graphrag}, modelnode-metrics, text width=38em]
                        ]
                    ]
                    [ 
                        Retrieval \\ Strategy, leaf-metrics, text width=6em
                        [
                            Multi-round Retrieval, leaf-metrics, text width=11em
                            [GoR~\cite{zhang2024graphofrecords}{,}
                            DialogGSR~\cite{park2024generative}{,}
                            Graph-CoT~\cite{jin2024graph}, modelnode-metrics, text width=38em]
                        ]
                        [
                            Post-retrieval, leaf-metrics, text width=11em
                            [KGR~\cite{guan2024mitigating}{,} Readi~\cite{cheng2024call}{,} GCR~\cite{luo2024graph}{,} 
                            CoK~\cite{wang2023boosting}, modelnode-metrics, text width=38em]
                        ]
                        [
                            Hybrid Retrieval, leaf-metrics, text width=11em
                            [StructRAG~\cite{li2024structrag}{,}
                            ToG 2.0~\cite{ma2024think-on-graph-2.0}{,} HYBGRAG~\cite{lee2024hybgrag}, modelnode-metrics, text width=38em]
                        ]
                    ]
                ]
                [
                    Knowledge\\Integration\\ (\S\ref{sec:integration}), leaf-methods,text width=6em
                    [ 
                        Fine-tuning, leaf-methods, text width=6em
                        [
                            Node-level Knowledge, leaf-methods, text width=11em
                            [SKETCH~\cite{anonymous2024large}{,}
                            GraphGPT~\cite{tang2023graphgpt}, modelnode-methods, text width=38em]
                        ]
                        [
                            Path-level Knowledge, leaf-methods, text width=11em
                            [RoG~\cite{luo2024reasoning}{,}
                            GLRec~\cite{wu2024exploring}{,}
                            KGTransformer~\cite{zhang2023structure}{,}
                            MuseGraph~\cite{tan2024musegraph}, modelnode-methods, text width=38em]
                        ]
                        [
                            Subgraph-level \\ Knowledge, leaf-methods, text width=11em
                            [GRAG~\cite{hu2024grag}{,} RHO~\cite{ji2023rho}{,}
                            GNP~\cite{tian2024graph}{,}
                            InstructGLM~\cite{ye2024language}{,}
                            LLAGA~\cite{chen2024llaga}, modelnode-methods, text width=38em]
                        ]
                    ]
                    [ 
                        In-context \\ Learning, leaf-methods, text width=6em
                        [
                            Graph-enhanced \\ Chain-of-Thought, leaf-methods, text width=11em
                            [Think-on-Graph~\cite{sun2024thinkongraph}{,}
                            Graph-CoT~\cite{jin2024graph}{,}
                            LARK~\cite{choudhary2023complex}{,}
                            Chain-of-Knowledge~\cite{li2024chainofknowledge}{,}
                            \\MindMap~\cite{wen2023mindmap}{,}
                            KnowledGPT~\cite{wang2023knowledgpt}, modelnode-methods, text width=38em]
                        ]
                        [
                            Collaborative Knowledge \\ Graph Refinement, leaf-methods, text width=11em
                            [KELP~\cite{liu2024knowledge}{,}
                            FMEA-KG~\cite{bahr2024knowledge}{,}
                            EtD~\cite{liu2024explore}{,} PoG~\cite{chen2024plan}{,} 
                            CogMG~\cite{zhou2024cogmg}, modelnode-methods, text width=38em]
                        ]
                    ]
                ]  
            ]
        \end{forest}
    }
     \vspace{-5mm}
    \caption{The taxonomy for existing GraphRAG methods in the survey.}
    \label{fig:taxonomy}
     \vspace{-5mm}
\end{figure*}
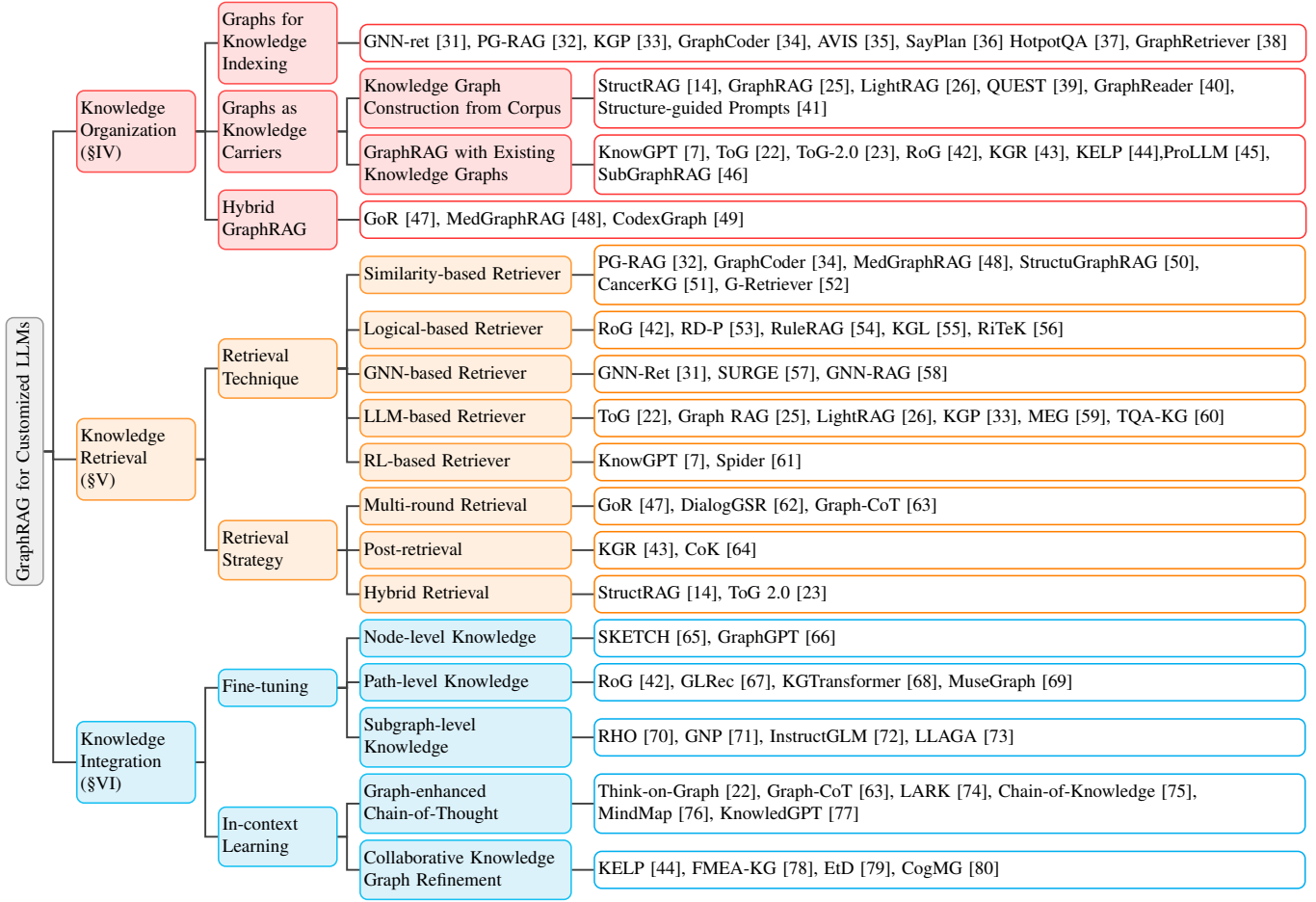

\subsubsection{Knowledge Integration}
The knowledge integration phase in RAG frameworks is crucial for synthesizing coherent and accurate responses based on both retrieved and inherent knowledge. At this stage, researchers utilize LLM to generate the output and employ several preprocessing and efficiency strategies to improve its quality and efficiency.

The quality of retrieved content can drastically affect generation, with irrelevant or misleading data potentially having deleterious effects. LeanContext~\cite{arefeen2024leancontext} addresses this issue by utilizing reinforcement learning to selectively choose sentences that are most pertinent to the query, effectively minimizing the context size and reducing computational costs. Similarly, SELF-RAG~\cite{asai2023self} introduces a self-reflection mechanism where the language model assesses its own generated and retrieved content, allowing for corrective measures to be implemented during the generation process. SKR~\cite{wang2023self} proposes a dynamic framework whereby LLMs can rely on their pre-trained knowledge for recognized queries, resorting to retrieval only when necessary. FILCO~\cite{wang2023learning} enhances this by training a context-filtering model that screens out irrelevant data, thereby mitigating hallucination risks. Recent research by Lyu et al.~\cite{lyu2023improving} introduces an evaluation method specifically designed to assess the importance of the retrieved content. Their findings suggest that by pruning or re-weighting parts of the retrieval corpus, RAG systems can enhance their performance without requiring additional training rounds. 
In scenarios where the retrieval corpus is limited, approaches such as Selfmem~\cite{cheng2024lift} construct a memory pool from LLM-generated results, using an iterative selection framework to enhance generative quality. SAIL~\cite{luo2023sail} forms an instruction-tuning dataset based on retrieval outcomes, fine-tuning LLMs to ground responses in reliable content while excluding distracting elements.


In conclusion, the generation stage within RAG focuses on bolstering the quality and efficiency of LLM-driven content generation. These methodologies continually evolve to maximize the potential of LLMs in generating context-aware, accurate, and efficient responses for RAG.

\subsection{Limitations of traditional RAG}
Although researchers have extensively explored traditional RAG, there are still some unresolved limitations due to the constraints of the data structure itself.

\subsubsection{Complex query understanding}
Traditional RAG faces significant challenges in precisely answering complex queries, mainly due to the intrinsic limitation of its knowledge organization (vector database). Given a query, these RAG methods only retrieve semantically similar chunks, within which the local contextual information may be insufficient to answer multi-hop questions.
This limitation becomes more pronounced as chunk granularity decreases, making it challenging to handle domain knowledge, which often requires multi-hop reasoning for effective understanding. 
Some traditional RAG methods have attempted to improve the complex query understanding, such as enhancing queries before retrieval~\cite{mao2020generation}, enhancing multiple query candidates for re-ranking~\cite{chuang2023expand}, and using related metadata to provide richer information~\cite{wang2024searchingforbestpractivesinrag}. However, these methods are still limited by the chunking style of knowledge organization and are challenging to capture multi-hop information in complex queries effectively.

\subsubsection{Distributed domain knowledge}
Domain-specific queries often involve jargon that requires contextual comprehension. However, domain knowledge is typically sparsely distributed across various documents and data sources. Although RAG uses chunking to divide documents into smaller pieces to manage this complexity and improve indexing efficiency, it sacrifices critical contextual information, significantly reducing retrieval accuracy and contextual understanding. Additionally, vector databases store the text chunks without a hierarchical organization of vague or abstract concepts, so it becomes difficult to resolve such queries, reducing the efficiency of utilizing expertise in specific domains. Some current traditional RAG works continuously align retrieved content with real-time generation states~\cite{jiang2024piperag}, and introduce external APIs to provide dynamic and rich auxiliary information~\cite{lazaridou2022internet}. While these approaches enhance RAG's understanding of dispersed domain knowledge, they still rely on the traditional chunking method for knowledge organization, which cannot fully address distributed domain knowledge. This limitation restricts their ability to construct robust connections between pieces of knowledge.

\subsubsection{Inherent constraints of LLMs}
Traditional RAG often uses the vector similarity-based retrieval module, which usually lacks effective filtering of the retrieved contents from vast knowledge bases, providing excessive but maybe unnecessary information. Considering the inherent constraints of LLMs, such as the fixed context window (typically 2K-32K tokens)~\cite{openai2023gpt4,anthropic2024claude} and the challenge of fully capturing long-range dependencies in complex documents, it is hard for LLMs in traditional RAG to capture the necessary information from the excessive retrieved contents. While scaling chunk granularity could alleviate these challenges, this approach significantly increases computational costs and response latency. Furthermore, indexing-based methods fail to effectively prune irrelevant information during retrieval, leading to the target knowledge being overwhelmed. These limitations present practical challenges for deploying RAG in resource-constrained environments where LLMs with smaller input context windows are preferred. Existing traditional RAG approaches have proposed methods such as recursive splits, sliding windows, and Small-to-Big~\cite{wang2024searchingforbestpractivesinrag,llamaindex_website} to optimize context length while preserving semantic integrity. They also assess and filter retrieved information to extract key insights~\cite{arefeen2024leancontext,asai2023self,wang2023learning,lyu2023improving}, and fine-tune LLMs to enhance their capability to capture crucial information from a vast amount of retrieved data~\cite{luo2023sail}. These approaches effectively reduce the volume of retrieved content, aiding the handling of LLMs' limited context windows, but they still fall short of capturing long-distance dependencies. This shortcoming arises because vector similarity-based retrieval makes it challenging to establish sufficiently clear connections between different pieces of retrieved content, limiting LLMs' understanding of long-distance dependencies.

\subsubsection{Efficiency and scalability}
Since large-scale knowledge sources often contain a significant amount of non-domain-specific information and domain-specific terms are usually sparsely distributed across diverse knowledge carriers, the retrieval module of RAG systems often needs to search through a vast amount of unstructured text to find relevant information, requiring considerable computational resources and time~\cite{edge2024local}. Although there are existing traditional RAG methods that propose preprocessing before employing retrieval methods to reduce costs~\cite{li2022decoupled,doostmohammadi2023surface,de2023pre}, or using innovative parallelism solutions to extract and maintain key information in intermediate processes to optimize resource use~\cite{liu2023tcra,xu2023recomp,jin2024ragcache}, they inevitably need to retrieve information from a large volume of unstructured text. This makes resource consumption a persisting issue for traditional RAG, affecting its scalability.

\begin{figure*}[htbp]
    \vspace{-3mm}
    \centering
    \includegraphics[width=\linewidth, trim=0cm 0cm 0cm 0cm,clip]{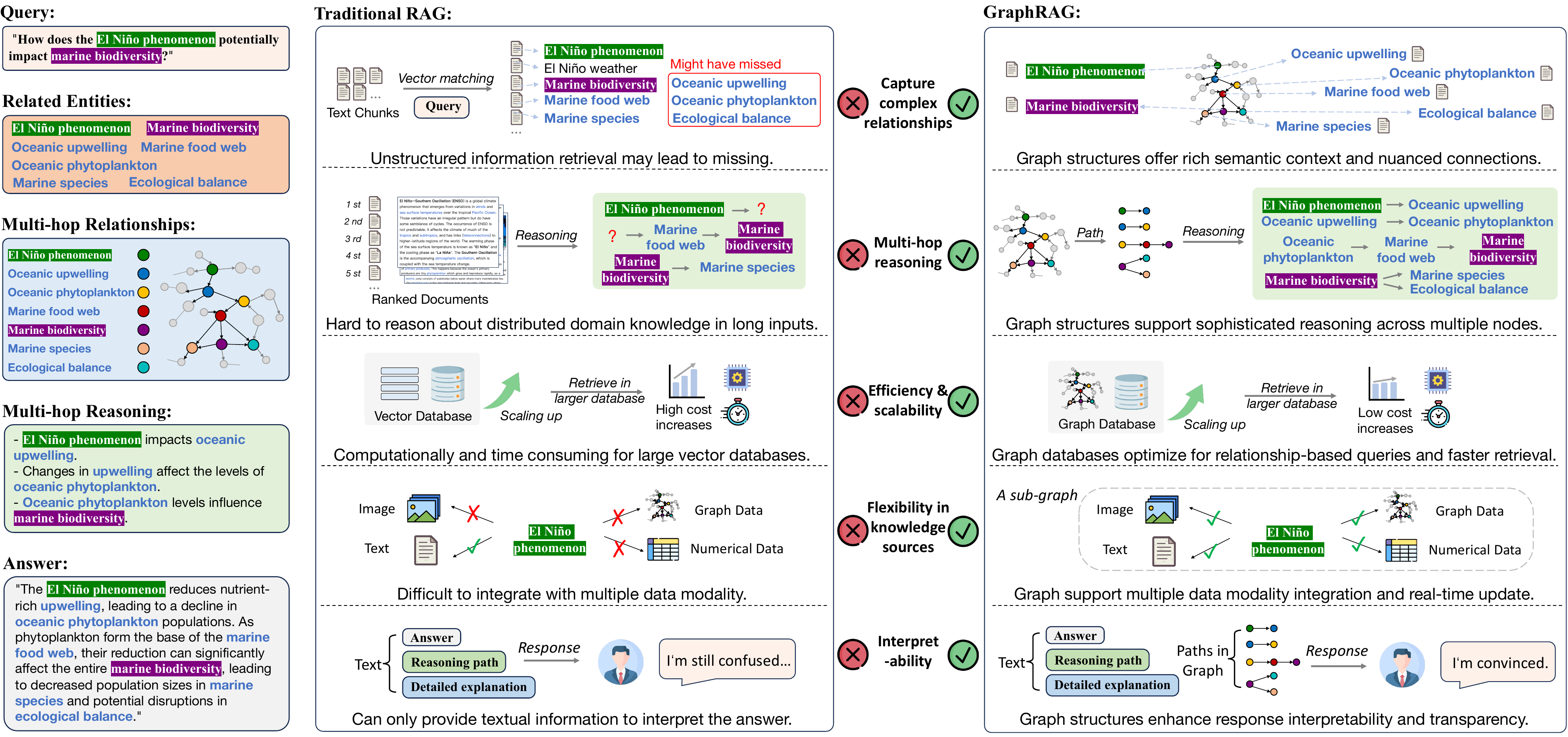}
    \vspace{-5mm}
    \caption{The illustration of the comparison between traditional RAG and GraphRAG.}
    \vspace{-5mm}
    \label{fig:RAGvsGraphRAG}
\end{figure*}

\subsection{GraphRAG}
To address the limitations of traditional RAG, a novel paradigm known as Graph Retrieval-Augmented Generation (GraphRAG) has been introduced. Leveraging structured knowledge, GraphRAG provides an efficient and accurate solution for organizing and retrieving information with structural databases, enhancing the performance and reliability of RAG systems.

\subsubsection{Definition of GraphRAG}
\label{def_graphrag}
GraphRAG can be formally defined as a subclass of RAG framework that  \textbf{leverage graph structure to organize and retrieve knowledge}. 
Unlike traditional RAG methods, which rely on vector databases for knowledge organization, GraphRAG employs structural databases where graphs are used to model dependencies among knowledge pieces.
This approach enhances the accuracy and efficiency of information retrieval, enabling more effective augmented generation for LLMs. Specifically, GraphRAG systems can utilize graphs as either the carrier of knowledge or the indexing tools for efficient retrieval from chunked textual data. 
By modeling dependencies between nodes, GraphRAG enables the discovery of related knowledge centered around a topic or anchor entity, ensuring comprehensive knowledge retrieval. Moreover, these connections support efficient search by navigating through relevant pathways and meanwhile pruning irrelevant information during the retrieval process.

\subsubsection{Workflow of GraphRAG}
Like traditional RAG, the workflow of GraphRAG can be divided into three key stages: knowledge organization, knowledge retrieval, and knowledge integration. However, due to the structured nature of GraphRAG, they have some special strategies compared to traditional RAG:

\noindent\textbf{Knowledge Organization} The knowledge organization stage structures external information using graph-based methods, either through explicit knowledge representation (graphs as knowledge carriers) or indexing mechanisms (graphs for knowledge indexing). 

\noindent\textbf{Knowledge Retrieval.} GraphRAG employs graph-based retrievers to retrieve relevant information based on the input query. This retriever not only considers the semantic similarity between the query and each text chunk but also the logical coherence between the query type and the retrieved subgraph.

\noindent\textbf{Knowledge Integration.} Once the relevant knowledge is retrieved, the GraphRAG model integrates it with the input query to generate the final output. The goal of the integration process is to seamlessly incorporate the retrieved knowledge into the generated text, thereby enhancing its quality and informativeness. A key design consideration is how to preserve the richness of the retrieved subgraph information in the final text-based prompt without introducing redundancy or misplacing emphasis on less critical aspects of the context.

\subsection{Traditional RAG vs GraphRAG}
GraphRAG provides several key advantages over traditional RAG systems, enhancing the capabilities of AI-driven information retrieval and generation. Below, we discuss these benefits and summarize the key points in Figure~\ref{fig:RAGvsGraphRAG}.

\subsubsection{Enhanced Knowledge Representation}
GraphRAG utilizes graph structures to represent knowledge, capturing complex relationships between entities and concepts. This approach allows for a more nuanced and contextual understanding of information compared to the flat document-based representation used in traditional RAG. The graph structure in GraphRAG can represent hierarchies, associations, and multi-hop relationships, providing a richer semantic context for queries and revealing non-obvious connections between different pieces of information. This capability can lead to new insights and discoveries, making GraphRAG particularly valuable in research and analysis applications.
 By representing multiple possible interpretations or relationships in the graph, GraphRAG can better handle ambiguous queries. It can explore different semantic paths and provide responses that account for various possibilities, offering a more nuanced understanding of complex topics.

\subsubsection{Flexibility in Knowledge Sources} GraphRAG systems can adapt to and integrate various knowledge sources, including structured databases, semi-structured data (like JSON or XML), and unstructured text. This versatility allows organizations to leverage their existing data infrastructure while benefiting from the advanced capabilities of GraphRAG. The system can connect data types, providing a unified view of an organization's knowledge. GraphRAG can incorporate different types of data (text, images, numerical data) into a single graph structure. This capability allows for more comprehensive knowledge representation and the ability to answer queries that span multiple data modalities.

\subsubsection{Efficiency and Scalability} GraphRAG systems built on fast graph databases can handle large-scale datasets efficiently. Graph databases are optimized for relationship-based queries, allowing for quick traversal of complex data structures. This efficiency translates to faster response times, especially for queries that require exploring multiple relationships. Research has shown that GraphRAG systems can generate LLM responses using 26\% to 97\% fewer tokens compared to traditional methods, indicating significant improvements in both speed and resource utilization.  GraphRAG systems can more easily accommodate updates to the knowledge base. New information can be added as nodes or edges in the graph without requiring a complete reindexing of the entire knowledge base. This feature allows for real-time updates and ensures that the system always has access to the most current information. The graph structure allows GraphRAG to consider the context of a query more effectively. Instead of relying solely on keyword matching or vector similarity, it can leverage the semantic relationships in the graph to retrieve more relevant information. This context awareness leads to more accurate and pertinent responses.
    
\subsubsection{Interpretability} The graph structure in GraphRAG allows for better visualization and understanding of how the system arrives at its responses. Users can trace the path of reasoning through the knowledge graph, seeing which entities and relationships were considered in formulating the answer. This transparency is crucial for building trust in AI systems and is especially valuable in fields like healthcare, finance, or legal applications where decision-making processes need to be auditable.

In conclusion, GraphRAG represents a significant advancement over traditional RAG systems, offering improved knowledge representation, reasoning capabilities, and efficiency. Its ability to capture and utilize complex relationships in data makes it a powerful tool for next-generation AI applications that require deep understanding and contextual awareness.


\section{Knowledge Organization}\label{sec:organization}
A major distinction between GraphRAG techniques and traditional RAG methods lies in their ability to leverage graph structures for efficient knowledge organization, enhancing the effectiveness of query responses. In this setup, the LLM functions as an intelligent agent, while the graph structures support its ability to organize and integrate information more comprehensively and precisely. In real-world applications, external knowledge sources may include extensive text corpora, document collections, search results, historical user data, or interaction logs. Properly organizing these sources is essential, as direct retrieval can be prone to irrelevant results and missing contexts, leading to information overload, knowledge conflicts, and compromised comprehension.

To address this, GraphRAG methods employ a two-step process: first constructing a graph structure to organize knowledge, then retrieving and integrating information relevant to the query. The organization of knowledge varies by task and source type, with three primary paradigms:
\textbf{\textit{ (1) Graphs for Knowledge Indexing}} for {\bf index-based} graphRAG: Text chunks are organized by symbolizing each chunk as a node within a graph, within which edges between nodes signify query-oriented relationships, making it easier to localize relevant knowledge. Considering knowledge hierarchy, further work builds multi-level index graphs, involving bottom-up knowledge summarization and top-down knowledge localization to enable coarse-to-fine knowledge indexing.
\textbf{ \textit{ (2) Graphs as Knowledge Carrier}} for {\bf knowledge-based} graphRAG: Here, summarized knowledge (e.g. atomic facts, community summarizations) is extracted from text chunks and integrated into a unified knowledge base, serving as a carrier of essential knowledge condensed from the raw corpora.
{\bf \textit{ (3) Hybrid GraphRAG}}: This approach combines both index graph and knowledge carrier functions, benefiting from the logical reasoning of knowledge graphs while preserving the detailed information in raw text chunks through indexing. The following sections detail each of these paradigms, explaining how each type of graph structure facilitates efficient and effective retrieval in diverse scenarios.

\subsection{Graphs for Knowledge Indexing}
Index-based GraphRAG methods utilize graph structures to index and retrieve relevant raw text chunks, which are then fed into LLMs for knowledge injection and contextual comprehension. These index graphs apply principles such as semantic similarities or domain-specific relations to effectively bridge connections among separate text passages. Compared to using graphs solely as knowledge carriers, this technique provides more informative answers by directly summarizing information from query-related raw text chunks.

The inherent challenges associated with the construction and
maintenance of these graphs raise several research questions, giving rise to several research questions that shape current research directions: {\bf (1) 
 Conciseness and Relevance}: Ensuring the constructed graph captures only relevant relationships without overloading it with unnecessary connections is a significant challenge, thus facilitating effective recalling of relevant text chunks without redundancies,
 {\bf (2) Consistency and Conflict Resolution}: Different data chunks may introduce conflicting information. It is crucial to resolve these conflicts and ensure that the graph remains consistent, reliable, and well-structured.
Researchers are actively exploring tailored index graph structures to address these challenges effectively.

For \textit{general-purpose} passages indexing, GNN-ret~\cite{li2024graph} creates a passage-level graph, connecting passages through structural and keyword similarities.  PG-RAG~\cite{liang2024empowering} treats LLMs as learners, forming a pseudo-graph by linking concise reading material summaries through common topics or complementary facts, generating an interconnected pseudo-graph database. KGP~\cite{wang2024knowledge} examines the logical associations of three major types of questions and develops three methods for constructing similarity-based index graphs. The first method uses TF-IDF to create lexical-similarity-based indexing graphs. The second employs KNN-ST/MDR~\cite{xiong2020mdr} to build semantic-similarity-based indexing graphs. The third utilizes TAGME~\cite{min2019knowledge} to construct indexing graphs based on shared entities.

A number of methods focus on \textit{domain-specific} index graph construction. In the code completion task, GraphCoder~\cite{liu2024graphcoder} utilizes a code context graph (CCG) that captures control flow and data dependencies between code statements, aiding in retrieving related code snippets for completion. AVIS~\cite{hu2024avis} generates a transition graph based on user decisions, mapping states, and permissible actions. For robot task planning, SayPlan~\cite{rana2023sayplan} employs 3D scene graph representations, enabling LLMs to perform semantic searches for planning purposes. These specialized methods tailor index graphs to meet the unique demands of various domain-specific tasks.

In summary, index graph-based methods have evolved from general-purpose passage and entity retrieval strategies to sophisticated pseudo-graphs for coherent information organization and, finally, to highly specialized graphs for domain-specific applications. Each development has addressed unique challenges, from organizing large text corpora and contextual summaries to leveraging inductive bias for enhanced domain-specific tasks like code completion, decision-making, and robotic planning. This evolution underscores the adaptability of graph-based structures in meeting the diverse and growing demands of LLM-driven applications across multiple domains.

\subsection{Graphs as Knowledge Carriers}
The paradigm of representing knowledge explicitly through graph structures is increasingly recognized for its effectiveness, offering several notable advantages: \textbf{(1) Efficient retrieval of query-related knowledge facts.} The explicit structure of knowledge graphs~\cite{pan2024unifying} facilitates logic-guided chain retrieval, efficiently identifying missing facts while pruning the search space through reasoning paths; \textbf{(2) Coherent multi-step reasoning over long spans.} Previous studies have revealed the importance of the reasoning steps, i.e., lengthening the reasoning steps in prompts brought significant reasoning performance enhancement~\cite{jin-etal-2024-impact}. Knowledge graphs excel in maintaining logical coherence for multi-step reasoning, where natural language often struggles. By leveraging planning algorithms, they identify optimal reasoning paths, enhancing both interpretability and clarity in handling complex information; \textbf{(3) Linguistic diversity.} Natural language often presents challenges with varying terminologies for the same entities or relations. KGs address this by enabling efficient entity and relation resolution through off-the-shelf graph algorithms, streamlining the integration of diverse data sources.

Research in this area can be categorized into two main directions based on the source of the knowledge graphs: \textit{GraphRAG with KGs constructed from corpus}, which transforms unstructured text into structured graph representations, and \textit{GraphRAG with existing KGs}, which focus on how to effectively leverage well-established KGs. Below we present the details of each direction and discuss the limitations of them.

\subsubsection{Knowledge Graph Construction from Corpus}
This research direction aims to transform unstructured text corpora into structured KGs, enabling efficient and precise information retrieval. It heavily relies on \textit{Open Information Extraction} techniques, which automate the domain-independent extraction of relational facts from large corpora. These techniques include both traditional OIE approaches~\cite{yates2007textrunner, angeli2015standfordoie,li2022constructing,kolluru2020openie6}
and advanced OIE methods based on LLMs~\cite{cohen2023crawling,chen2024sac-kg}. However, constructing effective KGs from domain-specific corpora remains challenging due to variations in ontology across domains. A key objective in this area is to incorporate inductive biases to design graph structures that capture essential domain-specific information while minimizing unnecessary complexity.

In scenarios where the query prompt consists of short passages and no external corpus is provided, Structure-guided Prompts~\cite{cheng2024structure} generate logically coherent responses by categorizing reasoning tasks into six types based on logical dependencies and organizing the passage's information using tailored graph structures specific to each type.
Other studies have focused on scaling language models to handle multi-document inputs through dynamically constructed knowledge graphs.
For instance, \cite{fan2019using} transformed sentences into relational triples via Coreference Resolution and Open Information Extraction, subsequently reducing graph redundancy by merging nodes and edges based on TF-IDF overlap.
Similarly, AutoKG~\cite{chen2023autokg} constructs a weighted knowledge graph where nodes represent keywords extracted from the original text, and edges, evaluated through graph Laplacian learning, represent relationships with assigned weights.
To offer more informative knowledge augmentation, QUEST~\cite{quest-quasi}, GraphRAG~\cite{edge2024local}, and GraphReader~\cite{li2024graphreader} construct attributed KGs where nodes are enriched with side information. Specifically, QUEST emphasizes integrating entities, relations, types, and semantic alignments, while GraphRAG leverages an LLM to generate community summaries as enriched knowledge for related entities, subsequently using these summaries to produce partial answers for the entities.
Similarly, GraphReader employs an LLM to summarize text chunks into atomic facts and extract key elements (nouns, verbs, and adjectives). Each key element, combined with its relevant atomic facts, forms a node, with links established between nodes sharing elements in the same atom facts. To leverage LLMs for customer service, \cite{xu2024retrieval} propose using RAG to retrieve and integrate historical issues. They create an attributed knowledge graph where nodes represent past issues and edges denote their connections. Each node includes a tree structure of the issue's information, capturing both inter-issue and intra-issue relationships to improve question answering. Recent studies have extended graph structures to more sophisticated forms. For example, StructRAG~\cite{li2024structrag} considers five candidate structure types tailored to different knowledge-intensive tasks: tables for statistical tasks, graphs for long-chain tasks, algorithms for planning tasks, catalogs for summarizing tasks, and chunks for simple single-hop tasks. A DPO-based router is trained to determine the optimal structure for representing knowledge in each case.

\begin{table*}[t]
	\caption{Representative knowledge retrieval techniques and strategies used in different GraphRAG Systems.}
    \vspace{-2mm}
	\label{tab:retrieval}
	\resizebox{1.0\textwidth}{!}{
	\begin{tabular}{ccccccccc}

		\toprule
\multirow{2}{*}{} & \multirow{2}{*}{Catagory} & \multirow{2}{*}{GraphRAG Model}   & \multicolumn{2}{c}{Input} & \multicolumn{3}{c}{Implementation Details}   & \multirow{2}{*}{Output} \\ 
\cmidrule(lr){4-5} \cmidrule(lr){6-8}
& & &Query Side & Graph Side &Query/Graph Preprocess & Matching & Pruning method&\\\cmidrule(lr){1-9}

\multirow{23}{*}{\rotatebox{90}{Retrieval Techniques}} 
&\multirow{6}{*}{ Similarity-based}  &StruGraphRAG~\cite{zhu2024structugraphrag}   &query embedding  &entity embedding &BERT/BERT &similarity calculation &\xmark &literal context \\
&  &CancerKG~\cite{gubanov2024cancerkg}  & keywords & entity  &NA/NA &TF-IDF   &\xmark &tabluar results \\
&  &G-Retriever~\cite{he2024g}  &query embedding & entity/relation embedding  &   SentenceBert/SentenceBert & k-nearest neighbors &PCST &subgraph \\

& &PG-RAG~\cite{liang2024empowering}  & query embedding  & Pseudo-Graph  &SentenceBert/SentenceBert   & depth-first search  & PGR & literal context, path \\
&  &GraphCoder~\cite{liu2024graphcoder}  & query slice  & CCG node  & CodeBERT/CodeBERT &Jaccard index  & subgraph edit distance  & code snippet \\
& & MedGraphRAG~\cite{wu2024medical} & query   & entity description  &SentenceBert/SentenceBert   & top-down search  & bottom-up refine & triple \\

\cmidrule(lr){2-9}
&\multirow{5}{*}{Logical-based } &RoG~\cite{luo2024reasoning}  &query embedding & relation path & Llama2/path generation &breadth-first search &LLM agent &reasoning path \\
&  &RD-P~\cite{huang2024rd}  &query embedding  &topic entity   &RoBERTa/RoBERTa &path expansion &path discriminator &reasoning path \\
&  &RuleRAG~\cite{chen2024rulerag}  &query embedding  &rule bank   &Llama2/rule mining &similarity calculation&\xmark &literal context, path  \\
& &KGL~\cite{yang2024intelligent} & query embedding & clarification path & Roberta/path generation & similarity calculation &\xmark & literal context\\
& &RiTeK~\cite{huang2024ritek} & query embedding & triple graph & SentenceBERT/SentenceBERT & MCTS/R-MCTS & LLM agent & reasoning path\\

\cmidrule(lr){2-9}

&\multirow{3}{*}{GNN-based } &GNN-Ret~\cite{li2024graph}   & subquestions & graph feature& SentenceBERT/RGNN& DPR & \xmark &subgraph\\
&  &SURGE~\cite{kang2023knowledge}  &query embedding  &triple embedding  &T5-small/GNN  &similarity calculation & DHT &triple \\
&  &GNN-RAG~\cite{dong2024advanced}  &query embedding  &  graph feature  & NA/GNN&similarity calculation & \xmark & literal context \\
\cmidrule(lr){2-9}

&\multirow{6}{*}{LLM-based } &KGP~\cite{wang2024knowledge} &query embedding  &passage (node description)  & T5/T5 &TF-IDF &LLM agent  & literal context \\
& &Graph RAG~\cite{edge2024local} & query & community summary & \xmark/community detection & LLM agent & \xmark &literal context \\
&  &ToG~\cite{sun2024thinkongraph}  &query embedding  &relation path   &Llama2/relation exploration &beam search &LLM agent &reasoning path \\
& &LightRAG~\cite{guo2024lightrag} & keywords & entity, global keys &GPT-4o/GPT-4o &keyword-search &global keyword match &literal context\\
&  &MEG~\cite{cabello2024meg}  &query embedding  &graph feature   &SapBERT/SapBERT &token generation & disambiguation &literal context \\
& &TQA-KG~\cite{he2024enhancing}& keyword  & triple    &entity extract/\xmark &keyword-search &LLM agent &subgraph \\
\cmidrule(lr){2-9}

&\multirow{2}{*}{RL-based }& KnowGPT~\cite{Zhang-etal24KnowGPT} &query embedding  & entity/relation embedding &BERT & similarity calculation & RL agent & reasoning path \\
&  &Spider~\cite{shen2025insight}  &query embedding  & entity embedding &NA & similarity calculation & RL agent & subgraph \\

\midrule

\multirow{8}{*}{\rotatebox{90}{Retrieval Strategy}} 
&\multirow{3}{*}{Multi-round } &DialogGSR~\cite{park2024generative} &query embedding & linearized graph & T5/T5 &subgraph generation &\xmark & linearized subgraph \\
&  &Graph-CoT~\cite{jin2024graph}  &query embedding  & entity embedding  &Llama2/Llama2  &similarity calculation & LLM agent& subgraph \\
&  &GoR~\cite{zhang2024graphofrecords}  &query embedding  & entity embedding  &Mixtral-7B/GAT &similarity calculation &\xmark &literal context \\
\cmidrule(lr){2-9}
&\multirow{2}{*}{Post-retrieval } &CoK~\cite{wang2023boosting} &pseudo evidence  & triple  &NA/NA &F2-Verification & \xmark & final answer \\
&  &KGR~\cite{guan2024mitigating}  &claim  & entity, triple  &claim extraction/NA & claim verification  & \xmark &final answer \\
\cmidrule(lr){2-9}
&\multirow{2}{*}{Hybrid Retrieval } & StructRAG~\cite{li2024structrag} &query embedding & entity embedding & Qwen2/Qwen2   &similarity calculation &LLM agent & literal context \\
&  &ToG 2.0~\cite{ma2024think-on-graph-2.0}  & keyword & entity, relation & keyword extract/NA & similarity calculation & entity/relation prune & literal context, path \\

	\bottomrule
	\end{tabular}
    }
    \vspace{-5mm}
\end{table*}

\subsubsection{GraphRAG with existing KGs}

This line of research focuses on utilizing well-established knowledge graphs, which can be either domain-specific—such as Lynx~\cite{lynx} for multilingual compliance services in the legal domain, AceKG~\cite{acekg} for academic applications, SPOKE~\cite{SPOKE-biometicKG} for biomedical applications, STRING~\cite{szklarczyk2019string} for protein-protein interaction prediction, — or general-purpose, like DBpedia~\cite{auer2007dbpedia} and YAGO~\cite{yago4.5}. The central challenge is to develop planning algorithms that dynamically retrieve reasoning paths or subgraphs from these KGs, tailored to the specific characteristics of the queries and the underlying graph structures. Researchers are investigating techniques to efficiently navigate these large-scale graphs, aiming to extract meaningful insights while ensuring the retrieved information is both relevant and concise.

The retrieval of reasoning paths typically involves generating relation paths grounded in KGs as plans, which are then used to extract factual knowledge from the KGs in the form of reasoning chains (relation paths with anchored entities). 
ProLLM~\cite{jin2024prollm} applies graph-based RAG to predict protein-protein interactions, modeling signaling pathways as reasoning processes where biological signals pass from upstream proteins through intermediates to downstream proteins via the shortest path.
Approaches such as RoG~\cite{luo2024reasoning} and KGR~\cite{guan2024mitigating} directly utilize LLMs as planning agents to identify optimal plans. To address the low recall of reasoning paths caused by the incompleteness of KGs, ToG~\cite{sun2024thinkongraph} employs a beam search algorithm to dynamically explore the most probable relational triples, forming coherent reasoning chains. Its successor, ToG-2~\cite{ma2024think-on-graph-2.0}, enables joint retrieval from a KG and textual documents, through alternating between logic chain extensions on KGs that explore neighboring entities of current ones, and contextual knowledge expansion from retrieved relevant documents.
In the biomedical domain, KG-RAG~\cite{soman2023biomedical} focuses on retrieving accurate and trustworthy biomedical contexts. It optimizes the retrieval process by using techniques such as disease entity recognition, context pruning, and leveraging a biomedical KG~\cite{SPOKE-biometicKG}. 
To optimize the planning process, KnowGPT~\cite{Zhang-etal24KnowGPT} formulates the exploration of plans as a planning problem and solves it using deep reinforcement learning. Meanwhile, KELP~\cite{liu2024knowledge} adopts a simpler semantic-matching strategy, encoding both the question and candidate knowledge paths in an encoder fine-tuned on a latent semantic space, and selecting the knowledge paths with the smallest semantic distance. To balance precision and recall in reasoning chain retrieval, subgraphRAG~\cite{li2024simple} introduces a lightweight multilayer perceptron to extract subgraphs, discovering key facts within the KGs. This method allows the subgraph size to be adjusted, achieving an efficiency-effectiveness trade-off. More details about the retrieval techniques are presented in Section \ref{sec:retrieval}.

\subsubsection{Limitations}
Although using graphs as knowledge carriers is efficient and effective, these methods are limited in several aspects:
$(i)$ \textbf{Lack of high-quality KGs.} For directly using KGs as external knowledge bases, this line of research is constrained by the availability of high-quality KGs. Constructing KGs is resource-intensive, and although powerful KG refinement techniques~\cite{difflogic,LLM4EA,neusymea,Zhang-etal22Contrastive,dong2023active,zhang2023integrating}
have been developed to improve KG completeness, most publicly available KGs remain far from comprehensive. Furthermore, the absence of a unified ontology poses challenges to designing transferable planning algorithms. These limitations have motivated recent research to focus on constructing KGs from corpora. However, this line of work introduces additional challenges:
$(ii)$ \textbf{Trade-off between efficiency and effectiveness.} 
When constructing KGs from text corpora, the granularity of the extracted knowledge plays a crucial role in balancing efficiency and effectiveness. Preserving fine-grained information results in larger-scale KGs, which may hinder computational efficiency. Conversely, compact KGs may sacrifice important details, leading to potential information loss. This trade-off complicates the knowledge summarization process. Moreover, approaches like GraphRAG~\cite{edge2024local}, which leverage LLMs for knowledge summarization, suffer from the high token costs and unbearable high runtime overhead.


\subsection{Hybrid GraphRAG}

This paradigm utilizes graph structures both as carriers of knowledge and as indexing tools. A common approach involves constructing a graph that encapsulates key information from the original text, with each node linked to corresponding text chunks. These text chunks act as a complementary knowledge source, providing detailed contextual information.

To enhance long-context summarization, GoR~\cite{zhang2024graphofrecords} constructs a graph by creating nodes for text chunks and their LLM-generated responses to simulated queries, with edges linking the queries, the retrieved text chunks, and the responses. This structure captures complex correlations among the elements through query-specified indexing, offering a comprehensive representation for effective long-context global summarization.
In the medical domain, MedGraphRAG~\cite{wu2024medical} introduces a unique triple graph construction. This approach creates a triple-linked structure that connects user documents to credible medical sources and controlled vocabularies, utilizing medical domain relations. This facilitates evidence-based medical responses with enhanced safety and reliability when handling private medical data.
To enhance repo-level code tasks, CodexGraph~\cite{liu2024codexgraph} uses static analysis to construct a code graph that links source code symbols (e.g., MODULE, CLASS, FUNCTION) through edges denoting relationships (e.g., CONTAINS, INHERITS, USES). This graph acts as a knowledge carrier by implying the workflow or data flow of the code and serves as an index for the metadata (source code pieces) associated with each node.

\section{Knowledge Retrieval} \label{sec:retrieval}

Based on the well-built knowledge base with graph-based organization strategies, we need to conduct retrieval from the knowledge base. Specifically, given a user query and a graph knowledge base with dense information, retrieving factual information relevant to the given query from the knowledge base is very important in developing effective and efficient GraphRAG systems. This retrieval process forms the cornerstone of a robust GraphRAG architecture, directly impacting its overall performance and utility. 
In this section, we will introduce current knowledge retrieval methods for graph-based retrieval-augmented generation in detail.

\subsection{Overall Pipeline of Knowledge Retriever}

The knowledge retrieval process in GraphRAG focuses on extracting relevant background knowledge from a graph database in response to a given query. This process follows three distinct and sequential steps that transform raw graph data into usable, contextual knowledge.

\subsubsection{Query/Graph Preprocessing}
The preprocessing stage operates simultaneously on both the query and graph databases to prepare them for efficient retrieval. For query preprocessing, the system transforms the input question into a structured representation through vectorization or key term extraction~\cite{zhu2024structugraphrag,gubanov2024cancerkg,liu2024graphcoder}. These representations serve as search indices for subsequent retrieval operations. On the graph side, the graph database undergoes more comprehensive processing where pre-trained language models transform graph elements (entities, relations, and triples) into dense vector representations that serve as retrieval anchors~\cite{he2024g,wang2024knowledge,he2024enhancing,ma2024think-on-graph-2.0}. Additionally, some advanced retrieval models apply graph neural networks (GNNs) on the graph database to extract high-level structural features, while a few methods even adopt rule mining algorithms to generate rule banks as rich, searchable indexes of the graph knowledge~\cite{li2024graph,kang2023knowledge,sun2024thinkongraph,chen2024rulerag}.

\subsubsection{Matching}
The matching stage establishes connections between the preprocessed query and the indexed graph database. This process compares the query representations against the graph indices to identify relevant knowledge fragments. The matching algorithm considers both semantic similarity and structural relationships within the graph. Based on the matching score, the system retrieves connected components and subgraphs that demonstrate high relevance to the query, creating an initial set of candidate knowledge.

\subsubsection{Knowledge Pruning}
The pruning stage refines the initially retrieved knowledge to improve its quality and relevance. This refinement process addresses the common challenge of retrieving excessive or irrelevant information, particularly when dealing with complex queries or large graph databases. The pruning algorithm applies a series of refinement operations to consolidate and summarize the retrieved knowledge~\cite{he2024g,Zhang-etal24KnowGPT,liang2024empowering}. Specifically, the system first removes clearly irrelevant or noisy information. It then consolidates related knowledge fragments and generates concise summaries of complex graph knowledge. The purpose of knowledge summarization is to facilitate comprehensive knowledge synthesis, which is vital for the generation process of LLMs. By providing a distilled and focused summary, we enable LLMs to better understand the context and nuances of the information, leading to more accurate and meaningful responses.

\subsection{Retrieval Techniques}

In this subsection, we provide a comprehensive overview of the various retrieval techniques employed in the GraphRAG knowledge base. These techniques can be broadly categorized into six main classes, including string matching, semantic similarity, logical reasoning, LLM-based approaches, RL-based methods, and GNN-based models, as shown in Table~\ref{tab:retrieval}.

\subsubsection{Semantics Similarity-based Retriever}
The semantic similarity-based retriever primarily conducts appropriate answer retrieval by measuring the similarity between queries in discrete linguistic space or continuous vector space and the knowledge base~\cite{zhu2024structugraphrag,liang2024empowering,he2024g,gubanov2024cancerkg,liu2024graphcoder}.
(i) \textbf{Discrete space modeling}.
Discrete space modeling methods primarily leverage linguistic discrete statistical knowledge to directly model text strings, which is a kind of simple and straightforward retrieval technique that relies on the exact matching of query terms with entities and relations in the graph base. It employs algorithms like substring matching, regular expressions, and exact phrase matching to identify relevant nodes and edges~\cite{he2024enhancing,gao2023retrieval,gubanov2024cancerkg,ma2024think-on-graph-2.0}. Although straightforward to implement, string matching is limited by its inability to handle variations in terminology, synonyms, and contextual nuances. Despite these limitations, it serves as a foundational approach, especially useful in domains with well-defined vocabularies and minimal ambiguity.
(ii) \textbf{Embedding space modeling}.
Embedding space modeling techniques are employed to evaluate the contextual and conceptual connections between queries and the elements of a graph base to transcend the constraints of string matching~\cite{zhu2024structugraphrag,he2024g,liu2024graphcoder,liang2024empowering}. Utilizing approaches like pre-train language models and word embeddings, such as TF-IDF, Word2Vec, and GloVe~\cite{selva2021review,kenter2015short}, these methods capture the semantic ties that bind different components. By mapping both queries and graph entities into a continuous vector space, these sophisticated techniques enable the retrieval of semantically pertinent information, bridging gaps even in the absence of precise term matches. However, the limitation of semantics similarity-based models is that they neglect the full exploitation of graph structures resulting in a significant underutilization of the inherent advantages of graph bases.

\subsubsection{Logical Reasoning-based Retriever}
Logical rule-based retrieval employs symbolic reasoning to deduce and extract pertinent information from graph knowledge bases. This methodology encompasses the creation of logical rules and constraints that articulate the relationships and hierarchies intrinsic to the knowledge bases. Utilizing techniques such as rule mining~\cite{chen2024rulerag}, inductive logic programming~\cite{lee2024diakop}, and constraint satisfaction~\cite{luo2024reasoning}, this approach uncovers insights that may not be explicitly present in the data. By harnessing these logical inferences, GraphRAG systems are capable of retrieving information that is congruent with the underlying structure and semantics of the graph bases.

\subsubsection{GNN-based Retriever}
Graph neural networks (GNNs) have become a mainstream tool for graph modeling and mining, achieving significant results in tasks such as node classification~\cite{NIPS2017_graphsage,bei2023reinforcement,li2024gslb, liu2023rsc} and link prediction~\cite{zhang2018link, chen2024gcr,bei2023nonrec}.
The message-passing mechanism in GNNs operates by iteratively aggregating information from a node's neighbors to update its own representation, enabling the network to learn from the structure and features of the graph~\cite{NIPS2017_graphsage}. 
GNN-based knowledge retriever primarily utilizes graph neural networks to encode nodes within the constructed graph bases~\cite{li2024graph,park2024generative,kang2023knowledge,kang2023knowledge} . The localizing and positioning of knowledge mainly rely on the encoding similarity of the node representations that contain both sentiment meanings and structural relationship understandings. GNN-based retrievers require the training of a GNN encoder. Additionally, due to the lack of explicitly labeled data, the focus of training is on designing an appropriate loss function that enables the GNN to learn to accurately locate the target knowledge through representation encoding.

\subsubsection{LLM-based Retriever}
With the advent of powerful LLMs, retrieval techniques have increasingly incorporated these models to enhance the semantic understanding and contextual relevance of the retrieved information. LLM-based retrieval leverages the deep contextual embeddings and generative capabilities of LLMs to interpret queries, generate relevant queries for the KG, and even synthesize information from multiple sources within the knowledge bases~\cite{sun2024thinkongraph,guo2024lightrag,ma2024think-on-graph-2.0}. These models can perform tasks such as query expansion, contextual disambiguation, and the generation of tailored retrieval paths, thereby improving the overall effectiveness of the retrieval process.
Regarding the constructed graph base, the LLM-based knowledge retriever primarily focuses on leveraging the LLM to comprehend the graph and identify key subgraphs. Compared to methods based on semantic similarity, it provides more powerful text representations. Specifically, the LLM-based knowledge retriever capitalizes on the formidable comprehension and generation prowess of LLMs to interpret the query and extract or synthesize pertinent candidate instances. Utilizing similarity-based retrieval strategies, it calculates the affinity between these candidates and the nodes within the graph, thereby identifying the most pivotal subgraph segments. For instance, PG-RAG~\cite{liang2024empowering} initially employs LLMs to generate key points corresponding to the query. Subsequently, it identifies seed nodes within the graph by computing a similarity matrix. Starting from these seed nodes, it searches for candidate nodes and ascertains their contribution to the seed nodes through a contribution matrix. Ultimately, the path (fact trajectory) with the highest cumulative contribution is selected as the retrieved collection of node information. 
On the constructed hierarchical tree, Thought Graph~\cite{hsu2024thought} employs LLMs to recursively expand and vote on the generated biological process terms with the prompt, selecting the term set that most accurately describes the gene set.
HOLMES~\cite{panda2024holmes} begins the process by leveraging the named entities extracted from the question as the foundation for a breadth-first traversal of the entity-document graph. By harnessing the capabilities of an LLM, it extracts KG triples from the document nodes and elevates them to a hyper-relations KG. Subsequently, the hyper-relations KG undergoes pruning through the computation of cosine similarity between relational embeddings, thereby ensuring that only the information most pertinent to the question is preserved.
In general, the LLM-based retriever can guide the model through the use of prompts without model training. The need for retriever training arises when there is a requirement to utilize the retrieval results to guide the LLM in reflecting on or fine-tuning its existing knowledge to enhance its capability~\cite{luo2024reasoning}. Recent studies also reveal the capability of LLMs to achieve few-shot or zero-shot transferable performance in graph-related tasks~\cite{li2023survey-graphmeetllms,li2024zerog,li2024graphintelligence}, showcasing their potential for resource-constrained scenarios involving large-scale and diverse graphs.

\subsubsection{Reinforcement Learning-based Retriever}
Reinforcement learning (RL) provides an adaptive and dynamic strategy for retrieval within GraphRAG systems. By framing the retrieval process as a sequential decision-making challenge, RL-based methods enable an agent to learn and traverse the graph base in search of the most pertinent information, guided by environmental feedback. Advanced techniques like Deep Q-Networks~\cite{wang2024m}, Policy Gradients~\cite{kulkarni2024reinforcement}, and Actor-Critic~\cite{gupta2024stackfeed} methods are deployed to refine retrieval strategies progressively. This methodology endows the system with the capacity to enhance its retrieval performance continuously through active interaction and accrued experience. This process can be described as follows:
The relevant reasoning background lies in a question-specific subgraph $\mathcal{G}_{\text{sub}}$ that contains all the \textit{source} entities $\mathcal{Q}_s$, \textit{target} entities $\mathcal{Q}_t$, and their neighbors. An ideal subgraph $\mathcal{G}_{\text{sub}}$ is expected to have the following properties: 
(i) $\mathcal{G}_{\text{sub}}$ encompasses as many source and target entities as possible;
(ii) The entities, and relations within $\mathcal{G}_{\text{sub}}$ exhibit a strong relevance to question context;
(iii) $\mathcal{G}_{\text{sub}}$ is concise with little redundant information such that it can be fed into LLMs with limited lengths.
However, it is challenging to find such a $\mathcal{G}_{\text{sub}}$ since extracting a subgraph is NP-hard. To effectively and efficiently find a satisfactory $\mathcal{G}_{\text{sub}}$, some researchers develop tailored knowledge extraction methods that employ deep RL to sample reasoning chains in a trial-and-error fashion~\cite{Zhang-etal24KnowGPT,shen2025insight}.

\subsection{Retrieval Enhancement Strategies}
Based on the retrieval technologies, some strategies are designed to enhance the retrieved resources, aiming to improve the relevance of target queries. We will introduce insights into these strategies in this subsection.

\subsubsection{Multi-round Retrieval} Beyond static retrieval techniques, some GraphRAG systems incorporate contextual information and user feedback to refine and adapt the retrieval process dynamically, which we call multi-round retrieval. These approaches aim to gradually align the retriever more closely with specific use cases and evolving user intents~\cite{park2024generative,jin2024graph,zhang2024graphofrecords}.

\subsubsection{Post-retrieval} Traditionally, GraphRAG mainly adopted prior retrieval, which means the retrieval process is conducted before generation. Recently, a new type of post-retrieval strategy has received more and more attention~\cite{wang2023boosting,guan2024mitigating}. This kind of method conducts retrieval after the generation process. In this way, the retrieved results can be adopted to evaluate whether the generated answers are faithful and accurate. The evaluation can be used as the guidance for LLM rethinking and correcting.

\subsubsection{Hybrid Retrieval} Combining different forms of data for retrieval can integrate their strengths and complement their weaknesses. In GraphRAG systems, recent studies have adopted the use of both graphs and other data formats to collaboratively enhance the RAG process.
Specifically, to enhance the scope of the retrieval candidates, \textit{Graph+Vector RAG}~\cite{xu2024retrieval,sarmah2024hybridrag} combine the knowledge graph and the vector database as the retrieval sources and integrate the retrieved results for answer generation. For example, ToG-2~\cite{ma2024think-on-graph-2.0} performs joint retrieval on a knowledge graph and text documents. It begins with entities identified from the query and iteratively retrieves relevant knowledge.
Due to the static nature of knowledge graphs, to further ensure the timeliness and accuracy of the retrieved information, some efforts have begun to combine knowledge graphs with online web resources for joint retrieval, which can be summarized as \textit{Graph+Online Web Resource RAG}~\cite{xie2024weknow,mendes2024application,alhanahnah2024depesrag,dehghan2024ewek,zhao2024bridging,bayat2023fleek}.
Further, K2~\cite{deng2024k2} explores the LLM tuning with multiple retrieval sources, such as the textual corpus, self-instructed LLM knowledge, and online web resources.
In future works, hybrid RAG with multiple retrieval resources simultaneously can be a potential research direction for complex LLM system building under a wide range of knowledge.

\begin{table*}[t]
\centering
	\caption{Representative fine-tuning knowledge integration techniques used in GraphRAG systems.}
     \vspace{-2mm}
	\label{tab:integration}
	\begin{tabular}{ccccc}

		\toprule
		 Category & Model   & Input Format & Preprocess & Applied (L)LMs \\\hline                                         
\multirow{2}{*}{Node-level}  &
SKETCH~\cite{anonymous2024large}   & Node attribute & Concatenate Node Attribute &  LLAMA3-8B \\

& GraphGPT~\cite{tang2023graphgpt}  & Node attribute \& Graph Embeddings  & Learning Node Embeddings & Baichuan-7B   \\
\cmidrule(lr){1-5}
\multirow{4}{*}{Path-level}  
 &GLRec~\cite{wu2024exploring}   &  User-item description & Path Weight Calculation & LLAMA-7B  \\
 &KGTransformer~\cite{zhang2023structure}   &  Sampled Knowledge Graph Sequence & None & Transformers   \\ 

&MuseGraph~\cite{tan2024musegraph}   &  Node Path \& Node attribute & COT & LLAMA-7B \\  
 &RoG~\cite{luo2024reasoning}   &  Knowledge Graph Path & Instruction Generation & LLaMA2-Chat-7B \\  
\cmidrule(lr){1-5}
\multirow{4}{*}{Subgraph-level} &RHO~\cite{ji2023rho}   & KG Embeddings \& KG SubGraph  & None &  BART \\
&GNP~\cite{tian2024graph}  &  Graph \& Query Embeddings   & None  &  FLAN-T5-xlarge \\
&InstructGLM~\cite{ye2024language}   &  Node attribute \& Graph Embeddings  & Flatten subgraph to sequence &  LLAMA-7B \\
&LLAGA~\cite{chen2024llaga}  &  Node attribute \& Graph Embeddings  & Flatten subgraph to sequence  &  Vicuna-7B \\
	\bottomrule
	\end{tabular}
     \vspace{-5mm}
\end{table*}

\section{Knowledge Integration}\label{sec:integration}
The knowledge retrieval phase is crafted to gather pertinent documents from various external sources in alignment with a specified query. Following this, the integration phase focuses on seamlessly synthesizing documents obtained from knowledge retrieval into a cohesive prompt, simultaneously setting appropriate training goals for the purpose of optimization. Recognizing that LLMs form a sturdy and foundational framework, the exploration during the integration phase predominantly steers clear of modifying the internal structure of individual layers or necessitating a complete reinitialization of LLMs for training. In this section, we delve into the comprehensive pipeline, relevant technologies, and strategies for enhancing knowledge integration in different GraphRAG systems.

\subsection{Overall Pipeline of Knowledge Integration}
Integrating graph-retrieved knowledge into LLMs mainly includes two main ways: fine-tuning and in-context learning. The overall pipeline of these integration methods can be summarized as follows.

\textit{Fine-tuning}. To directly leverage information retrieved from graph searches to enhance open-source LLMs, fine-tuning offers a straightforward solution for the integration, such as LoRA-based tuning~\cite{hu2022lora,li2024loftq} and other data-efficient fine-tuning strategies~\cite{hu2023llm,lin2024data}. It injects the retrieved knowledge directly into the LLMs, focusing on graph-retrieved information at three knowledge levels: node-level knowledge, path-level knowledge, and subgraph-level knowledge for model tuning. In this way, graph information from different levels enhances the different capabilities of LLMs.

\textit{In-context Learning}. While numerous open-source LLMs have been released to date, many state-of-the-art LLMs remain closed-source in practice. 
The integration of closed-source LLMs is constrained since it is not feasible to jointly train or fine-tune closed-source LLMs in an end-to-end manner. 
Thus, in-context learning provides an indirect strategy for knowledge integration, which can be roughly decomposed into two steps: prompt format choice and LLM response optimization.
(i) Prompt format choice: 
The choice of prompt format is crucial for knowledge integration. This is because LLMs are highly sensitive to the prompt format. For example, the order of examples in in-context learning can lead to different responses~\cite{he2024does,arawjo2024chainforge}.
(ii) LLM response optimization: Then the appropriate prompt format joins the retrieved content and the questions as input to prompt the LLMs for response generation or further optimization.

\subsection{Integration Techniques}

\subsubsection{Fine-tuning Techniques}
The fine-tuning process, leveraging various graph information, can be delineated into three distinct categories based on the granularity of the input target:
(i) \textbf{Node-level Knowledge}: Focusing on individual nodes within the graph.
(ii) \textbf{Path-level Knowledge}: Concentrating on the connections and sequences between nodes.
(iii) \textbf{Subgraph-level Knowledge}: Considering larger structures composed of multiple nodes and their interconnections.
We will explore each of these aspects in detail.

\noindent\textbf{Fine-tuning with Node-level Knowledge.}
In many graph-based RAG systems, each node is linked to a document, such as an abstract in a citation network \cite{huang2024can}. Since domain-specific data is seldom present in pre-training corpora \cite{deng2024k2}, some studies employ instruction tuning to bolster domain-specific knowledge comprehension before proceeding with downstream task fine-tuning \cite{li2023chatdoctor, wu2024pmc}. A straightforward fine-tuning approach involves feeding node and neighboring text as contextual information into LLMs to aid in predictions \cite{fatehkia2024t, dehghan2024ewek,anonymous2024large}. Given that retrieved documents can be extensive, researchers can leverage LLMs to distill these texts into a single embedding \cite{lewis2020retrieval}. Despite the absence of pre-training data for out-of-vocabulary tokens, LLMs are capable of discerning the information within these embeddings in practice \cite{zhang2024text}. To augment the information within these embeddings, alignment techniques are integrated to fuse multi-modal data, such as images \cite{tang2023graphgpt}.

\begin{table*}[!t]
    \centering
    \caption{Representative in-context learning knowledge integration techniques used in GraphRAG systems.}
    \vspace{-2mm}
    \begin{tabular}{lcccccc}
    \toprule
    \textbf{Methods} & \textbf{Applied LLMs} & \textbf{Zero-shot} & \textbf{Few-shot} & \textbf{CoT} & \textbf{Graph Query} & \textbf{Release Time} \\
    \midrule
    LARK~\cite{choudhary2023complex} & LLaMA-2 & \xmark & \xmark & \cmark & \cmark & May-2023 \\
    Chain-of-Knowledge~\cite{li2024chainofknowledge} & GPT-3.5 &  \xmark & \cmark & \cmark &  \xmark & May-2023 \\
    Think-on-Graph~\cite{sun2024thinkongraph} & GPT-4, GPT-3.5, LLaMA-2 &  \xmark & \cmark & \cmark &  \xmark & Jul-2023 \\
    MindMap~\cite{wen2023mindmap} & GPT-3.5 & \cmark & \cmark & \cmark & \xmark & Aug-2023 \\
    KnowledGPT~\cite{wang2023knowledgpt} & GPT-4 & \cmark &  \xmark & \cmark & \cmark & Aug-2023 \\
    Align-NL2GQL~\cite{liang2024aligning} & Qwen, Baichuan2, ChatGLM & \xmark & \xmark & \xmark & \cmark & Feb-2024 \\
    Graph-CoT~\cite{jin2024graph} & LLaMA-2, Mixtral, GPT-3.5 & \xmark & \cmark & \cmark & \xmark & Apr-2024 \\
    GNN-RAG~\cite{mavromatis2024gnn} & LLaMA-2, GPT-3.5 & \cmark & \xmark & \xmark & \xmark & May-2024 \\
    FMEA-KG~\cite{bahr2024knowledge} & GPT-4 & \xmark & \xmark & \xmark & \cmark & Jun-2024 \\
    KELP~\cite{liu2024knowledge} & GPT-3.5 & \xmark & \cmark & \xmark & \xmark & Jun-2024 \\
    CogMG~\cite{zhou2024cogmg} & Qwen & \xmark & \cmark & \xmark & \xmark & Jun-2024 \\
    EtD~\cite{liu2024explore} & LLaMA-2 & \cmark & \xmark & \xmark & \xmark & Jun-2024 \\
    \bottomrule
    \end{tabular}
    \label{tab:methods}
    \vspace{-5mm}
\end{table*}

\noindent\textbf{Fine-tuning with Path-level Knowledge.}
Linguistic tasks often involve intricate reasoning and require a clear understanding of factual relationships \cite{bianchini2024enhancing}. Utilizing knowledge graph paths, LLMs are guided through the transitory relationships and entities, thereby enhancing their reasoning capabilities with evidence-based support \cite{shu2024knowledge}. These paths can either be the most direct routes from question entities to answer entities or be mined using graph retrieval models \cite{mavromatis2024gnn} or heuristic methods \cite{tan2024musegraph}. They can function as both input and output, but when multiple paths exist between two nodes, it is crucial to filter out noisy paths while preserving the relationships within the knowledge graph \cite{wu2024exploring}. To maintain the integrity of entity representations and their relationships along the paths, certain approaches focus on using these paths as training objectives, predicting nodes and relations along a path between two nodes \cite{zhang2023structure} or even across multiple paths \cite{luo2024reasoning}. This enables LLMs to engage in edge-level reasoning and produce reliable outputs.

\noindent\textbf{Fine-tuning with Subgraph-level Knowledge.}
A subgraph represents a segment of a graph, encompassing a subset of nodes and edges from the original structure. Unlike the linear topology of path data, subgraph data exhibits a more complex, irregular topology \cite{tang2023graphgpt}. This complexity arises because subgraphs can contain a multitude of connections beyond those found in paths, allowing them to capture intricate relationships between nodes and posing a greater challenge for LLMs to learn \cite{chen2024graphwiz,chen2024llaga}. One straightforward approach is to employ a graph encoder to condense subgraph-level information into a readout embedding \cite{he2024g, tian2024graph}. Alternatively, some researchers transform graph data into sequences, leveraging LLMs' inherent strength in processing sequential data, such as executable graph database queries \cite{luo2023chatkbqa}. However, these methods often overlook the rich textual content within subgraphs and fail to make LLMs cognizant of the underlying graph structure.
To address this, efforts are divided: some focus on adapting the transformer architecture to better handle structured data \cite{ji2023rho, yan2024enhancing, yuan2024gnnavi}, while others incorporate descriptions of nodes and edges directly into the prompt \cite{ye2024language, chenllaga}. However, there still remain challenges for existing methods. The former risks losing knowledge acquired during pre-training due to architectural alterations, while the latter may struggle with dense graphs featuring a large number of nodes and edges.

\subsubsection{In-context Learning Techniques}
Recent research on knowledge integration through in-context learning (ICL) has introduced sophisticated frameworks that can be broadly divided into two stages: 
(i) \textbf{Graph-enhanced Chain-of-Thought}: This approach leverages graph structures to enhance the reasoning process within LLMs.
(ii) \textbf{Collaborative Knowledge Graph Refinement}: This method highlights the distinct techniques used in optimizing LLM responses through collaborative refinement of well-built knowledge graphs.

\noindent\textbf{Graph-enhanced Chain-of-Thought.}
Since Chain-of-Thought (CoT) has demonstrated effective performance improvement and scalability~\cite{zhang2023automatic}, CoT-based prompting techniques on knowledge integration also elicit strong potential in complex multi-hop reasoning. The Graph-enhanced CoT, such as Reasoning on Graphs~\cite{luo2024reasoning}, emphasizes chain reasoning based on the interaction between graphs and LLMs.
Think-on-Graph~\cite{sun2024thinkongraph,ma2024think-on-graph-2.0} proposes a reasoning framework based on the KG-LLMs interaction; LLMs are required to retrieve the relevant sub-graph of the proposed keyword first, then deduce the final result through chain reasoning. Similarly, in Graph CoT~\cite{jin2024graph}, given a question, LLM reasons the key component in the external graph that is required for the final answer, then uses a multi-round graph execution and interaction to finish the chain reasoning that concludes the final answer. Chain-of-Knowledge~\cite{li2024chainofknowledge}, firstly constructs exemplars that answer the given question based on evidence triples retrieved from the knowledge base, then utilizes faithfulness and factuality verification to check the reliability of the chain prediction. Program-of-Thought (PoT)~\cite{wang2023plan,bi2024program} is also a well-designed prompting technique that extends CoT prompting into coding scenarios. Following this, the KnowledGPT~\cite{wang2023knowledgpt} framework uses PoT to generate the research language of knowledge bases to retrieve the answer step by step.
The CoT-based methods achieve promising progress in knowledge integration for GraphRAG, significantly improving the reasoning capability on complex questions. However, error accumulation is still a severe problem~\cite{ma2021markov,bai2024transformers}. To address this, research has been conducted to diversify the reasoning paths that LLMs can take to reach the same conclusion~\cite{li2024different,xie2022an}, which can help mitigate the effects of error propagation. Besides, the integration of visual in-context learning for Large Vision-Language Models is being explored to enhance cross-modal interactions and alignment~\cite{zhou2024visual,zhang2023makes}, which may provide further insights into managing error accumulation during reasoning.

\noindent\textbf{Collaborative Knowledge Graph Refinement}. 
Instead of a well-designed prompting technique utilized to generate a better prompt or elicit LLMs for better reasoning, refining the LLM’s original response based on the factual knowledge in knowledge graphs is also an effective method to prevent LLMs from hallucination scenarios. 
The timeliness and accuracy of knowledge graphs are crucial, as they greatly influence the quality of augmented generation~\cite{jin2024graph}. Maintaining and refining the content in knowledge graphs through feedback in LLM generation is an important strategy in GraphRAG.
Representatively, KG-based Retrofitting~\cite{guan2024mitigating} is a framework that combines LLMs with KGs to mitigate hallucination during the reasoning process. It retrofits the initial draft responses of LLMs based on the factual knowledge stored in KGs with an autonomous knowledge verifying and refining procedure. Similarly, KELP~\cite{liu2024knowledge} refines a trainable encoder for path selection based on the LLM’s response. The Explore-then-Determine framework~\cite{liu2024explore} uses the integration of LLMs and KGs to determine the final answer. CogMG~\cite{zhou2024cogmg} has designed a framework for collaborative augmentation between LLMs and KGs, leveraging KGs to augment LLMs in generations, explicitly targeting incomplete knowledge. LLMs are then required to identify and decompose required knowledge triples that are not present in the KG, enriching them and aligning updates with real-world demands.
The collaboration between KGs and LLMs aligns factual knowledge to ensure the accuracy of generation and the up-to-date quality of KGs. However, ensuring the correctness of the intermediate process is challenging, and the accuracy of the refining process should also be evaluated with tailored metrics~\cite{xu2024search}.

\subsection{Integration Enhancement Strategies.}

\subsubsection{Training with other domain-specific models}
Training with domain-specific models is a critical approach to enhance the capabilities of LLMs in handling multi-modal information representations and to improve their performance on specific tasks.
(i) Enhancing Multi-modal Information Processing: 
Integrating multi-modal information representations, such as image representations learned by vision-language models~\cite{geng2022improving}, into LLMs, is a significant advancement. These representations not only reduce input lengths but also encapsulate rich information from specific inputs. 
However, they also pose challenges for LLMs, as they may be treated as out-of-vocabulary tokens, hindering the full exploitation of the original multi-modal information. Furthermore, recent efforts demonstrate the large number of additional tokens, especially in the extremely sparse case where each additional token is only paired with a few training samples, can cause LLMs to perform even worse over the small LMs. 
To address this, recent efforts have focused on fine-tuning LLMs with domain-specific models to increase their generalization capabilities across different domains~\cite{tariq2024domain}.
(ii) Fine-tuning LLM with Domain-specific Tasks:
Fine-tuning an LLM involves using a pre-trained model and refining its weights by training it on a small set of annotated data with a slow learning rate. This principle allows the language model to adopt new knowledge from the data while retaining its initial learnings~\cite{shen2024tag}. Transfer learning, a technique that allows a pre-trained model to apply its knowledge to a new task, is particularly useful when sufficient datasets for fine-tuning are not available~\cite{jeong2024fine}. For instance, MedPaLM~\cite{singhal2023large} is a domain-specific model built upon PaLM, which demonstrated exceptional performance in complex tasks by using prompting with annotated questions and answers.

\subsubsection{Multi-round Integration}
Integrating knowledge through a single round of interaction often falls short of providing satisfactory answers. To this end, researchers have been exploring multi-round integration techniques that enhance the quality of responses: (i)Enhancing Retrieved Content Quality: 
One of the primary reasons LLMs may fail to answer questions effectively in a single round is the inconsistency in the quality of retrieved content.
Multi-round integration allows for iterative refinement of the retrieved information. For instance, the IM-RAG approach~\cite{yang2024rag,tanwar2023multilingual} integrates retrieval systems with LLMs to support multi-round RAG through learning Inner Monologues, which are akin to the human inner voice that narrates one's thoughts.
(ii) Handling Multi-hop Complex Reasoning: Complex questions often require multi-hop reasoning. Multi-round integration enables the generation of intermediate reasoning steps, leading to more accurate reasoning results. The study~\cite{yang2024large} comprehensively analyzed the latent multi-hop reasoning capabilities of LLMs. 
They found that while LLMs can exhibit multi-hop reasoning, their performance is significantly influenced by the structure of the prompt and the relational information within.
(iii) Aligning Output with Target Labels: LLMs' output may not always meet specific requirements, such as aligning with target labels. To overcome this, multi-round integration can help in aligning the model's output with the desired outcomes. The X-InSTA method~\cite{tanwar2023multilingual} proposes a cross-lingual in-context source-target alignment strategy that aligns prompt examples in a cross-lingual scenario.
(iii) Multi-round Interaction for Quality Assurance: Multi-round integration can better ensure the stability and quality of the answers.
For example, AgentPS~\cite{liu2024agentps} integrates agentic process supervision into LLMs via multi-round question answering during fine-tuning, highlighting the effectiveness of integrating process supervision and structured sequential reasoning for multimodal content quality assurance.

\section{Conclusion}
LLMs have demonstrated impressive capabilities in natural language processing, achieving remarkable performance across a wide range of tasks. However, their effectiveness diminishes significantly in specialized domains, especially when faced with knowledge-intensive tasks requiring domain expertise that have almost never appeared in the pre-training corpus. 
RAG has emerged as a promising solution that enhances LLMs with external knowledge bases to improve their domain-specific capabilities. However, traditional RAG systems encounter several critical challenges in customizing LLMs for specific domains, including (i) efficient processing of extensive document collections, (ii) effective integration of knowledge dispersed across multiple sources, and (iii) maintenance of contextual coherence during information retrieval and generation. 
These limitations become particularly critical in applications requiring complex reasoning, where successful inference depends on both comprehensive knowledge integration and deep contextual understanding. To address these challenges, GraphRAG emerges as a pioneering approach that enriches LLMs with well-organized background knowledge. This survey provides a comprehensive analysis of GraphRAG, detailing its taxonomy, mechanisms, challenges, and future research directions. Besides, we also provide valuable resources for practitioners to deploy GraphRAG in production environments.

\bibliographystyle{IEEEtran}

\bibliography{main}

\appendix


\subsection{Open-Source Projects and Applications}\label{sec:opensource}

The advent of GraphRAG has opened new avenues for enhancing information retrieval and generation processes through the integration of graph-based methodologies. As researchers and developers explore the potential of GraphRAG, a variety of open-source projects and applications have emerged, demonstrating its versatility across different domains. This section delves into the real-world implementation of GraphRAG systems, including benchmark datasets,  that serve as evaluation standards and downstream tasks that gauge their effectiveness in real-world scenarios. Additionally, it highlights notable open-source projects that exemplify the implementation of GraphRAG principles, paving the way for innovative applications in areas such as chatbots, knowledge management, and research development.

\subsubsection{Benchmark Datasets}

We categorize representative datasets based on their complexity and specific characteristics:
\paragraph{Simple Question Answering}
\begin{itemize}
    \item \textbf{SimpleQuestion}~\cite{bordes2015large} consists of 100k questions constructed from the Freebase knowledge graph. The dataset emphasizes basic evidence retrieval without requiring complex reasoning chains. Its straightforward nature makes it ideal for evaluating baseline performance and validating fundamental retrieval mechanisms in KGQA systems. The questions are designed to test direct fact extraction, making it a valuable resource for assessing retrieval accuracy.
    \item \textbf{WebQ}~\cite{berant2013semantic} comprises 4,737 questions, split into 3,098 training and 1,639 testing examples. The questions were collected through Google Suggest API, capturing natural user queries. Each question is annotated with SPARQL queries, enabling systematic evaluation of semantic parsing capabilities. The dataset leverages the Freebase knowledge graph, providing a realistic test bed for question-answering systems dealing with real-world information needs.
\end{itemize}
\paragraph{Multi-hop Reasoning}
\begin{itemize}
    \item \textbf{CWQ}~\cite{talmor2018web} extends WebQSP to 34,689 questions requiring sophisticated reasoning patterns. The dataset is carefully balanced across four question types: composition (45\%) involving multiple-step reasoning chains, conjunction (45\%) combining multiple constraints, comparative (5\%) handling numerical or temporal comparisons, and superlative (5\%) dealing with maximum or minimum value queries. Questions require up to 4-hop reasoning paths, making it a comprehensive benchmark for evaluating complex reasoning capabilities over the Freebase knowledge graph.
    \item \textbf{MetaQA}~\cite{zhang2018variational} specializes in movie domain knowledge, containing over 400k questions structured around a knowledge graph with 135k triples, 43k entities, and 9 relations. The dataset is organized into progressive difficulty levels based on reasoning hop requirements. Each question includes detailed annotations of head entities, answers, and the entities involved in the reasoning path, providing a controlled environment for evaluating multi-hop reasoning capabilities.
\end{itemize}

\paragraph{Large-scale Complex QA}
\begin{itemize}
    \item \textbf{LC-QuAD}~\cite{trivedi2017lc} contains 5,000 question-SPARQL pairs based on DBpedia 2016. The dataset supports both public endpoint access and local query execution, making it versatile for different evaluation setups. It focuses on testing systems' ability to generate and execute complex SPARQL queries accurately.
    \item \textbf{KQAPro}~\cite{cao2020kqa} represents the largest KGQA dataset with 94,376 training and 11,797 validation/test examples. Built on a dense subset of Wikidata, it features multiple inference types and logical operations including unions and intersections. Each example provides both SPARQL queries and KoPL logical forms, with entities and predicates represented in their natural form rather than ID-based notation.
\end{itemize}
\begin{table*}[t]
	\caption{Open-source Projects and Applications.}
	\label{app_tab:code_link}
    \vspace{-3mm}
	\resizebox{1.0\textwidth}{!}{
	\begin{tabular}{lcccc}

		\toprule
		Model & Domain & Task & Dataset & Background Knowledge                                                  \\\hline
Graph RAG~\cite{edge2024local} &General Domain  &Graph Construction, QA &- & Podcast Transcripts, News Articles \\
ToG~\cite{sun2024thinkongraph} &General Domain &QA &WebQSP, CWQ &Freebase \\
ToG 2.0~\cite{ma2024think-on-graph-2.0} &General Domain &QA & WebQSP, CWQ &Freebase \\
RoG~\cite{luo2024reasoning} &General Domain &QA &WebQSP, CWQ &Freebase \\
KnowGPT~\cite{Zhang-etal24KnowGPT}& General Domain & QA &CommenconseQA, OpenbookQA &ConceptNet\\ 
SubgraphRAG~\cite{li2024simple}&General Domain &QA & WebQSP, CWQ& Freebase\\
NQ-RAG~\cite{dong2024advanced}&General Domain &Graph Construction, QA &NQ &Wikipedia \\
LBR-GNN~\cite{yang2024language}&General Domain &QA &CommonsenseQA,OpenbookQA &Wikipedia, ConceptNet \\
SG-RAG~\cite{saleh2024sg}&General Domain &QA &MetaQA, LC-QuAD, ComplexWebQuestions &Freebase\\
StructRAG~\cite{li2024structrag}&General Domain &QA &Loong, Podcast Transcripts &-\\
R-DP~\cite{huang2024rd}& General Domain &QA &WebQSP, CWQ & Freebase \\
GAIL~\cite{zhang2024gail}& General Domain &QA &WebQSP, CWQ, GrailQA & Freebase \\
KG-RAG~\cite{soman2023biomedical}&Biomedical&QA &MCQ &SPOKE \\
MEG~\cite{cabello2024meg} &Medicial &QA &MedQA, PubMedQA, MedMCQA &UMLS \\

DialogGSR~\cite{park2024generative}&Medical &Dialog Generation &OpenDialKG,KOMODIS &Freebase, IMDb \\
ReTek~\cite{huang2024ritek}&Medical &Graph Construction, QA &ReTek &PharmKG, ADInt \\
KG4Diagnosis~\cite{zuo2024kg4diagnosis}&Medical &Graph Construction, Dialog Generation &- &SNOMED-CT, UMLS\\
CancerKG~\cite{gubanov2024cancerkg}&Medical &QA &CancerKG &PrimeKG, PubMed\\
MedGraphRAG~\cite{wu2024medical}&Healthcare, Medical & QA &MultiMedQA, DiverseHealth &MedC-K, UMLS \\
AGENTiGraph~\cite{zhao2024agentigraph}&Legislation, Healthcare  &Graph Construction, QA &- & UK Legislation, MMedC \\
KGL~\cite{yang2024intelligent}&Water Conservancy &Graph Construction, QA &KGQS &Special Report \\
SURGE~\cite{kang2023knowledge}& Movie &Dialog Generation &OpendialKG, KOMODIS  &Freebase, IMDb \\
TQA-KG~\cite{he2024enhancing}&Education &Graph Construction, QA &TQA-KG &CK12-QA, AI2D \\
GraphFusion~\cite{yang2024graphusion}&Scientific Research &Graph Construction, QA &TutorQA &NLP KG, TutorialBank, NLP-Papers \\
StructuGraphRAG~\cite{zhu2024structugraphrag}&Scientific Research &Graph Construction, QA &NSDUH & Codebook \\
Soccer-GraphRAG~\cite{sepasdar2024soccer}&Soccer &Graph Construction, QA &Soccer KG &SoccerNet-Echoes \\
LightRAG~\cite{guo2024lightrag}&Cross Domain &Graph Construction, QA &- &UltraDomain \\
PG-RAG~\cite{liang2024empowering}&Cross Domain &Graph Construction, QA &Mindmap &CRUD \\
DiaKoP~\cite{lee2024diakop}&Cross Domain &Dialog Generation &ConvQuestions &Wikidata \\
Graph-CoT~\cite{jin2024graph}&Cross Domain &QA &- &-\\
G-Retriever~\cite{he2024g}&Mutiple Domain &Graph Construction, QA &GraphQA &ExplaGraphs, SceneGraphs, WebQSP \\
	\bottomrule
	\end{tabular}
    \vspace{-5mm}
    }
\end{table*}

\paragraph{Domain-specific QA}
\begin{itemize}
    \item \textbf{Mintaka}~\cite{sen2022mintaka} contains 20k questions across 8 different languages, focusing on logical refinement and answer revision capabilities. The dataset emphasizes cross-lingual reasoning and the ability to systematically improve answers based on available evidence.
    \item \textbf{FACTKG}~\cite{kim2023factkg} encompasses 108,000 claims requiring validation against DBpedia. The binary classification task tests systems' ability to verify factual accuracy using knowledge graph information, providing a unique perspective on knowledge validation capabilities.
    \item \textbf{WebQSP}~\cite{yih2016value} features 4,737 natural language questions operating over a vast knowledge graph containing 164.6M facts and 24.9M entities. The questions are distributed across different reasoning types: 30\% require two-hop aggregation, 7\% involve constraint reasoning, and 63\% need single-fact retrieval.
    \item \textbf{GrailQA}~\cite{gu2021beyond} focuses on complex question answering over knowledge bases, incorporating advanced reasoning patterns. The dataset challenges systems to demonstrate sophisticated query understanding and execution capabilities, pushing the boundaries of KGQA system development.
    \item \textbf{TutorQA}~\cite{yang2024graphusion} is a dataset specifically designed for research in educational question answering and tutoring systems. It consists of questions and answers extracted from real-world tutoring interactions. 
    \item \textbf{CRUD}~\cite{lyu2024crud} is constructed by collecting recent high-quality news articles from major Chinese news websites, which provides a fresh and challenging benchmark for RAG systems.
    \item \textbf{UltraDomain}~\cite{qian2024memorag} is a benchmark for evaluating RAG systems on complex queries. It includes 428 college textbooks across 18 domains, such as agriculture, humanities, and computer science, covering topics like machine learning and big data. 
\end{itemize}

\paragraph{Tailored GraphRAG Benchmarks}
\begin{itemize}
     \item \textbf{GraphRAG-Bench}~\cite{xiang2025use} is a tailored benchmark for evaluating GraphRAG systems. It features a comprehensive dataset with tasks of increasing difficulty, covering fact retrieval, complex reasoning, contextual summarization, and creative generation, and a systematic evaluation across the entire pipeline, from graph construction and knowledge retrieval to final generation.
\end{itemize}

\subsubsection{GraphRAG Applications}

GraphRAG systems have emerged as a powerful framework for real-world applications, offering robust solutions to challenges across diverse domains by leveraging graph-based representations of knowledge. These systems excel in tasks like QA, dialog generation, and other natural language processing tasks, where they integrate structured and unstructured data to enable advanced reasoning and contextual understanding. Their versatility stems from their ability to utilize a wide range of datasets and background knowledge, adapting to the specific needs of various fields.

In the \textbf{general domain}, GraphRAG models such as Graph RAG~\cite{edge2024local}, SubgraphRAG~\cite{li2024simple}, and StructRAG~\cite{li2024structrag} are designed to handle tasks that require broad contextual understanding. These models utilize extensive knowledge sources, including Freebase, Wikipedia, ConceptNet, and even unstructured data such as podcast transcripts and news articles. By constructing graphs that represent relationships between entities, these systems are capable of answering complex queries that go beyond simple fact retrieval. For example, GraphRAG models can synthesize information from different domains to address multifaceted questions, making them highly effective for general-purpose knowledge exploration and reasoning.

In the \textbf{biomedical} and \textbf{medical domains}, GraphRAG systems have proven particularly impactful. Models like KG-RAG~\cite{soman2023biomedical}, MED-RAG~\cite{zhao2025medrag}, MEG~\cite{cabello2024meg}, and MedGraphRAG~\cite{wu2024medical} are tailored to handle the intricate and high-stakes nature of medical data. They rely on specialized datasets such as PubMedQA, MedQA, and UMLS, which provide rich sources of structured medical knowledge. By constructing graphs that capture relationships between medical concepts, these models can assist in a range of tasks, including medical QA, diagnostic support, and the integration of healthcare data. For instance, KG-RAG leverages knowledge graphs to link symptoms, diseases, and treatments, enabling healthcare professionals to access accurate and interpretable insights. Similarly, MedGraphRAG addresses the need for personalized healthcare solutions by integrating multimodal data from sources like MultiMedQA and DiverseHealth.

In \textbf{legislation area}, GraphRAG systems like AGENTiGraph~\cite{zhao2024agentigraph} bridge the gap between regulatory requirements and practical applications. AGENTiGraph focuses on integrating legislative and healthcare-related knowledge graphs to address domain-specific QA tasks. By utilizing datasets such as UK legislation and MMedC (medical corpora), this model enables the navigation of complex regulatory environments, such as understanding compliance requirements in healthcare or analyzing the implications of new policies. This integration of structured legislative data and healthcare knowledge allows AGENTiGraph to support decision-making processes in areas where legal and medical considerations intersect.

In the realm of \textbf{education} and \textbf{scientific research}, GraphRAG systems like TQA-KG~\cite{he2024enhancing} and GraphFusion~\cite{yang2024graphusion} play a critical role in facilitating knowledge dissemination and academic inquiry. These models are designed to process datasets such as CK12-QA, AI2D, and TutorQA, which provide educational content in structured formats. By constructing knowledge graphs that represent interconnected concepts, these systems enable students, educators, and researchers to access relevant information efficiently. For example, TQA-KG can assist in answering educational queries by linking concepts across subjects, while GraphFusion integrates scientific literature and tutorial resources to support researchers in navigating complex academic topics. For scientific research, StructGraphRAG~\cite{zhu2024structugraphrag} organizes and processes research papers, integrating data from scientific corpora to create graphs that highlight relationships between key findings, methodologies, and datasets. This capability aids researchers in identifying trends and synthesizing new insights.

Beyond these domains, GraphRAG systems demonstrate their adaptability to specialized fields. In \textbf{water conservancy}, models like KGL~\cite{yang2024intelligent} utilize datasets such as KGQS to create domain-specific knowledge graphs that facilitate the management and analysis of water resources. These graphs enable stakeholders to identify relationships between hydrological data, infrastructure, and policies, improving decision-making processes. In the domain of \textbf{sports} analytics, Soccer-GraphRAG~\cite{sepasdar2024soccer} applies graph-based reasoning to soccer data, utilizing datasets like SoccerNet-Echoes to analyze player performance, match statistics, and team dynamics. This capability supports applications such as player scouting and strategy development. In the \textbf{entertainment industry}, systems like SURGE~\cite{kang2023knowledge} are used to construct knowledge graphs centered on movie-related data. These graphs enhance applications such as content recommendation, audience analysis, and storytelling, enabling deeper engagement with viewers. 

Overall, GraphRAG systems demonstrate unparalleled versatility by adapting their graph-based frameworks to meet the unique demands of various real-world domains. Their ability to integrate structured and unstructured data into cohesive knowledge representations enables them to address complex problems across general and highly specialized fields. By constructing graph bases tailored to specific applications, these systems contribute to advancements in healthcare, education, legislation, scientific research, sports analytics, and other specific domains. Their impact lies in their ability to provide interpretable, context-aware insights that empower decision-making and drive innovation.

\subsubsection{Open-source Project}

GraphRAG has inspired a range of open-source projects that explore its principles and adapt its framework to various use cases. These implementations highlight the versatility and growing interest in combining knowledge graphs with RAG techniques to enhance LLMs.

\textbf{Microsoft GraphRAG}~\footnote{\url{https://github.com/microsoft/graphrag.git}} serves as the foundational framework that combines knowledge graphs with RAG to enhance the performance of LLMs. It constructs a knowledge graph from private datasets, leveraging graph machine learning to enrich the query process through prompt augmentation at runtime. By doing so, Microsoft GraphRAG achieves remarkable improvements in answering complex queries that require detailed reasoning or domain-specific mastery. Its ability to outperform traditional methods demonstrates its effectiveness in handling private datasets where accuracy and contextual understanding are critical.

Building upon this foundation, a variety of open-source projects have emerged to explore and adapt GraphRAG's principles for diverse needs. One such project is \textbf{Nano-GraphRAG}~\footnote{\url{https://github.com/gusye1234/nano-graphrag.git}}, a lightweight and highly customizable implementation designed with simplicity in mind. Featuring a concise codebase of approximately 1,100 lines, Nano-GraphRAG is accessible to developers aiming to prototype or deploy graph-enhanced LLM systems without the overhead of a full-scale solution. It supports multiple graph storage systems, such as Neo4j, and integrates seamlessly with vector databases like Milvus and Faiss, offering flexibility for a range of applications.

\textbf{Azure GraphRAG}~\footnote{\url{https://github.com/Azure-Samples/graphrag-accelerator.git}} extends the concept by offering a solution accelerator built on the graphrag Python package. This project provides API endpoints hosted on Azure, allowing users to trigger indexing pipelines and query knowledge graphs seamlessly. Azure GraphRAG focuses on demonstrating how knowledge graph memory structures can enhance LLM outputs in a hosted service environment. While not an official Microsoft product, it serves as a practical toolkit for developers and researchers to experiment with and deploy GraphRAG-based solutions, bridging the gap between research concepts and real-world applications.

\textbf{Fast GraphRAG}~\footnote{\url{https://github.com/circlemind-ai/fast-graphrag.git}} takes a performance-oriented approach to the GraphRAG concept. By leveraging asynchronous operations and parallelized graph querying, it prioritizes speed and efficiency, making it well-suited for real-time applications that demand rapid data retrieval and reasoning. This implementation highlights the importance of optimization for time-sensitive domains, where traditional graph traversal methods might fall short.

In contrast, \textbf{LightRAG}~\footnote{\url{https://github.com/HKUDS/LightRAG.git}} focuses on reducing computational complexity and resource requirements, providing a streamlined solution for environments with limited infrastructure. Its lightweight architecture makes it ideal for deployment on edge devices or within constrained systems, such as IoT networks or small-scale reasoning platforms.

Another notable adaptation is \textbf{Medical GraphRAG}~\footnote{\url{https://github.com/SuperMedIntel/Medical-Graph-RAG.git}}, a domain-specific variant tailored to the healthcare industry. This implementation integrates medical knowledge graphs, such as SNOMED and ICD, to provide contextually accurate and clinically relevant outputs. By bridging the gap between medical terminology and generative AI capabilities, Medical Graph RAG facilitates applications like clinical decision support and patient education, where precision and reliability are paramount.

These open-source projects, along with the original Microsoft GraphRAG, collectively demonstrate the adaptability and growing importance of integrating structured knowledge graphs with LLMs. They highlight the potential of this approach to create intelligent, context-aware systems that address complex challenges across a wide array of industries and domains.

\subsection{Limitations and Future Opportunities}\label{sec:future}\label{sec:future}
In this section, we systematically analyze the critical limitations of existing GraphRAG regarding knowledge quality, knowledge conflict, data privacy, and efficiency issues and discuss potential research directions for practical advancements.

\subsubsection{Knowledge Quality} 
The effectiveness of GraphRAG models fundamentally depends on the quality of the external knowledge, necessitating the development of sophisticated mechanisms for knowledge engineering. This encompasses advanced techniques for (i) systematic knowledge organization, (ii) automated quality refinement, and (iii) intelligent knowledge base expansion. First, knowledge organization demands more expressive graph structures that capture complex semantic relationships, temporal dynamics, and hierarchical dependencies through hybrid neuro-symbolic approaches and advanced embedding techniques. Second, knowledge refinement requires automated quality assurance frameworks that leverage cross-validation, statistical analysis, and machine learning to identify inconsistencies, remove redundancies, and validate factual accuracy. Third, knowledge expansion is critical in enhancing GraphRAG systems' practical effectiveness through continuous knowledge enrichment and adaptation. The practical implementation of knowledge expansion typically combines multiple approaches: automated web crawling for public information updates, API integrations with authoritative databases, expert feedback loops for validation, and machine learning models for relationship inference. This multi-faceted approach ensures robust and reliable knowledge growth while maintaining data quality and relevance. The seamless integration of these components, supported by a scalable infrastructure for dynamic updates and version control, will be crucial for developing robust GraphRAG systems across various applications.

Beyond traditional text-based knowledge, integrating multi-modal information, including images and videos, offers promising opportunities to enrich the external knowledge databases with more comprehensive domain expertise and conceptual understanding. Equally crucial is developing robust knowledge validation frameworks to ensure reliability and maintain data integrity across different modalities. These frameworks should incorporate advanced techniques from anomaly detection, knowledge base completion, and alignment to systematically identify and correct errors, address information gaps, and maintain consistency across the integrated knowledge base. Such comprehensive approaches to knowledge quality assurance will be fundamental in enhancing the reliability and effectiveness of GraphRAG systems across diverse applications.

\subsubsection{Knowledge Conflict}
Integrating multiple knowledge sources in GraphRAG models introduces significant challenges in managing conflicting information and maintaining knowledge consistency. As such, another critical research priority is developing sophisticated techniques for knowledge reconciliation, conflict resolution, and truth discovery. These approaches should incorporate advanced algorithms capable of identifying contradictory statements, evaluating source reliability, and determining the most probable accurate information based on available evidence and contextual facts. Beyond conflict resolution, ensuring seamless alignment between external knowledge and LLM-generated outputs presents another fundamental challenge. This alignment requires innovative methods for knowledge distillation, fine-tuning, and cross-modal integration to harmonize structured knowledge with the LLM's learned representations. Incorporating uncertainty modeling and probabilistic reasoning frameworks offers a promising direction for handling ambiguous or conflicting information more effectively. By associating knowledge with confidence scores and probability distributions, GraphRAG systems can make more nuanced decisions and generate outputs that accurately reflect the inherent uncertainty in the integrated knowledge. This comprehensive approach to knowledge consistency, combining conflict resolution, alignment techniques, and uncertainty modeling, will be essential for developing more robust and reliable GraphRAG systems capable of handling real-world information complexity.

\subsubsection{Data Privacy}
The integration of external knowledge in GraphRAG systems raises critical privacy concerns that demand sophisticated technical solutions and robust governance frameworks. Privacy-preserving knowledge integration and retrieval represent critical challenges requiring advanced cryptographic approaches, including secure multi-party computation, homomorphic encryption, and differential privacy mechanisms. These techniques enable GraphRAG systems to leverage sensitive information from knowledge graphs while maintaining strict privacy guarantees for individuals and organizations. Beyond technical solutions, developing comprehensive data governance frameworks becomes essential for responsible deployment. Such frameworks should establish clear standards for access control, data handling protocols, and ethical guidelines that align with evolving privacy regulations while maintaining system utility. This dual focus on technical privacy preservation and governance frameworks will be crucial in building trustworthy GraphRAG systems that balance the competing demands of knowledge accessibility and privacy protection, ultimately enabling broader adoption across sensitive domains such as healthcare, finance, and personal information management.

\subsubsection{Efficiency}
The practical deployment of GraphRAG systems in real-world applications hinges critically on model efficiency. Future work should investigate techniques for optimizing the knowledge retrieval and integration processes to reduce computational overhead and memory requirements. This may involve exploring knowledge distillation, pruning, and compression methods to create more compact and efficient representations of domain knowledge. Developing scalable and efficient algorithms for subgraph matching, graph traversal, and reasoning is another important direction for future work. By leveraging techniques from graph theory, database systems, and parallel computing, researchers can enable faster inference times and real-time generation of responses in GraphRAG models. Furthermore, investigating hardware acceleration techniques, such as utilizing GPUs and TPUs, can significantly boost the performance of GraphRAG models. Exploiting the parallelism and computational power of specialized hardware can enable the efficient processing of large-scale knowledge graphs and support the integration of extensive knowledge into language models. Through advancements in these research directions, future research can significantly improve the practical applicability of GraphRAG systems by reducing computational overhead, memory requirements, and response latency while maintaining or enhancing system performance.

\subsection{Discussion on GraphRAG}\label{sec:discussion}
\subsubsection{Knowledge Retrieval in GraphRAG}
The true magic of graph-based knowledge resources lies in their supernatural ability to connect data while being flexible and scalable during retrieval. Graphs can seamlessly integrate various data sources, including structured and unstructured data. This fusion creates a unified view of information, often revealing hidden connections and patterns. Advanced graph knowledge bases incorporate ontologies and semantic schemas, providing formal definitions for concepts and their interrelationships within a domain. While traditional retrieval methods are like tourists asking random pedestrians for directions, GraphRAG is the mayor who knows every nook and cranny of the city. It doesn't just look at individual buildings (or pieces of information); it sees the entire cityscape, considering the intricate network of streets, neighborhoods, and communities. This bird's-eye view allows GraphRAG to navigate these streets with the efficiency of a seasoned cab driver who knows all the shortcuts.
Here is a vivid example: Let’s consider GraphRAG to be a sophisticated urban planning and information retrieval system for the knowledge base aka city. This system operates in two main phases: planning and retrieval.

In the planning phase, the city planners (LLMs) first divide the urban landscape (source documents) into manageable districts (concepts). They then survey each district, identifying key landmarks and connections (element instances) - think of this as mapping out important buildings, parks, and the roads connecting them. These individual elements are then summarized into concise descriptions (element summaries), like creating brief profiles of each neighborhood landmark. Next, using advanced urban analysis tools (community detection algorithms like Leiden), the planners group these elements into natural communities, much like identifying distinct neighborhoods or boroughs within the city. Finally, they craft detailed "neighborhood reports" (community summaries) for each of these communities, providing a rich, multi-layered guidebook to the entire city. This indexing phase creates a comprehensive, hierarchical understanding of the urban landscape, from individual buildings to entire districts.

After the planning, we obtain a high-quality graph knowledge base. Then, when it comes to retrieval, GraphRAG shines in handling citywide inquiries. Instead of dispatching surveyors to random locations, it consults the pre-prepared neighborhood reports. Each community contributes its local perspective to the broader question, like local town halls holding simultaneous meetings (community summaries to community answers). These local insights are then synthesized into a comprehensive city report (global answer), much like an urban planner combining feedback from various boroughs to understand citywide trends. This approach proved particularly effective for "global" questions about the entire urban landscape, outperforming traditional methods in both the breadth and diversity of insights offered.

\subsubsection{Knowledge Integration in GraphRAG}

The knowledge integration allows the users to inject domain-specific knowledge into LLMs. When integrating, the following aspects need to be considered for the design:

\noindent\textbf{Computational Overhead} Incorporating retrieved text into the input consumes a portion of the available input length. The computational overhead may increase further due to the complexity of handling graph-structured data from graph-retrieved information. GraphRAG combines well-built graphs with LLMs, which requires additional processing to retrieve and manage graph elements such as nodes, triples, and paths. This added complexity can lead to much higher costs.
Given that prompt processing time scales quadratically with the length of the prompt, large retrieved texts can significantly amplify the computational overhead for LLMs, especially in downstream applications like billion-scale recommender systems~\cite{wang2024llms,kim2024large}. To mitigate this, domain-specific knowledge should be permanently integrated into LLMs through fine-tuning, enhancing their responses, and tailoring them to specific applications. This approach could lead to specialized LLMs that reduce reliance on extensive text retrieval from external sources, thereby improving response speed and reducing both time and financial costs~\cite{xia2024understanding,sun2024cebench}. For instance, deploying GPT-4 for pilot-scale customer service could incur costs in the thousands of dollars, while fine-tuning a large language model might require hundreds of gigabytes of memory. The practical deployment of LLMs in real-world scenarios is thus heavily influenced by these cost considerations. 

\noindent\textbf{Graph-structured Input} LLMs are not inherently equipped to process graph-structured data. The common practice is to convert graph data into natural language descriptions, such as explaining the connections between nodes~\cite{liu2024can,huang2024can-tmlr}. However, these descriptions alone are insufficient for LLMs to grasp the complex geometric properties of graphs. Graph-structured data is a fundamental input format for numerous tasks, including node classification and link prediction in text-attributed networks. The success of LLMs in graph-related tasks depends on their capacity to understand and process graph-structured inputs. As graph reasoning tasks grow in complexity, such as calculating the minimum cut problem for a randomly generated graph, the importance of accurate input modeling becomes even more critical.

\noindent\textbf{Large-scale Graph} The scalability of models is crucial for dealing with real-world graphs \cite{wan2023scalable}. Handling graph-structured input becomes particularly challenging with large graphs, where the number of edges and nodes may surpass the input length limitations of LLMs. Therefore, efficiently managing large-scale graph problems within the constraints of LLMs is a significant area of research \cite{bi2024lpnl}. Advances in this field could enable LLMs to handle more complex and larger graphs, which is essential for real-world applications such as social network analysis, bioinformatics, and knowledge graph construction.

\subsection{Traditional RAG Pipeline}\label{sec:ragpipeline}
A RAG framework begins by retrieving relevant information from pre-constructed external knowledge bases based on the query. This information is then used to prompt LLMs, guiding them in constructing credible reasoning chains. As a result, RAG enables LLMs to generate more substantiated and accurate content, effectively minimizing hallucinations and inconsistencies. The traditional RAG pipeline typically comprises three core components: knowledge organization, knowledge retrieval, and knowledge integration.

\subsubsection{Knowledge organization}
In traditional RAG systems, knowledge organization involves structuring and preparing external knowledge repositories to facilitate rapid and relevant retrieval when provided with a query. A common strategy is to split the large-scale text corpus into manageable chunks. These chunks are then transformed into embeddings using an embedding model, where the embeddings serve as keys of original text chunks in a vector database~\cite{borgeaud2022improving,izacard2023atlas,jiang2023active}. This setup enables efficient look-up operations and retrieval of relevant content via distance-based search in the semantic space.

As a crucial step in the pre-retrieval process, several methods~\cite{gao2023retrieval,wang2024searchingforbestpractivesinrag} have been proposed to optimize knowledge organization, focusing on two main aspects: granularity optimization and indexing optimization. Granularity optimization aims to balance relevance and efficiency, as coarse-grained units provide richer context but risk redundancy and distraction, while fine-grained units may lack semantic integrity and increase the retrieval burden~\cite{yu2023chain,zhong2024mixofgranularity}. To control granularity, chunking strategies are employed to split documents into chunks based on token limits. Methods such as recursive splits, sliding windows, and Small-to-Big~\cite{wang2024searchingforbestpractivesinrag,llamaindex_website} strive to maintain semantic completeness while optimizing context length. Indexing optimization seeks to improve the structure and quality of content for retrieval. Metadata-addition techniques, which attach chunk text with metadata like titles, timestamps, categories, and keywords, enable filtering and re-ranking operations during the post-retrieval process~\cite{wang2024searchingforbestpractivesinrag}. Another type of technique is hierarchical indexing, which organizes files into parent-child relationships with summaries at each node, facilitating faster and more efficient data traversal while reducing retrieval errors~\cite{wang2024corag}. Such tree-like indexing methods represent early attempts at structured knowledge organization and have inspired successors to harness the power of graph structures for knowledge organization, i.e., GraphRAG.

In summary, knowledge organization is foundational to GraphRAG. By carefully constructing the knowledge resources, RAG systems can ensure the reliability of retrieved content.

\subsubsection{Knowledge retrieval}
The knowledge retrieval stage encompasses various methods and strategies designed to efficiently access and retrieve the necessary knowledge from pre-organized repositories, ensuring the selection of relevant information that can enhance the quality of generated outputs.

Current RAG works usually involve retrieval methods such as k-nearest neighbor retrieval (KNN), term frequency-inverse document frequency (TF-IDF), and best matching 25 (BM25) to retrieve the relevant content.
RETRO~\cite{borgeaud2022improving} employs KNN to extract approximate relevant neighbors from the conducted key-value database by calculating the L2 distance. RETROprompt~\cite{chen2022decoupling} extends this approach with a few-shot knowledge store, tailoring it for more advanced prompts.

In addition, some specialized techniques are used prior to the retrieval method to improve the accuracy and efficiency of the retrieval. To capture multifaceted aspects of the query, GAR~\cite{mao2020generation} introduces diverse context generation, enriching the initial query with additional contexts before applying BM25 retrieval. Enhancing this framework, EAR~\cite{chuang2023expand} implements a re-ranking process that selects the optimal candidate from multiple expanded queries to improve the retrieval accuracy. Furthermore, in tackling the computational challenges associated with exact retrieval methods like BM25, Doostmohammadi et al.~\cite{doostmohammadi2023surface} propose a hybrid approach to identify approximate neighbors using sentence transformers for representation and then apply BM25 for re-ranking, effectively balancing accuracy with computational efficiency on large-scale retrieval tasks.

Retrieval model training is another crucial aspect, where recent methodologies have explored the use of self-supervised techniques. For instance, REPLUG~\cite{shi2023replug} demonstrates that labels generated by a frozen language model can be leveraged to directly supervise the training of retrieval models. This approach enhances retrieval quality without requiring additional manually annotated datasets. ATLAS~\cite{izacard2023atlas} fine-tunes both the retriever and the LLMs in tandem. Furthermore, frameworks like FLARE~\cite{jiang2023active} dynamically balance retrieval with generation needs through an active RAG framework that intelligently determines retrieval timings based on the generation progression.

Moreover, the use of external APIs helps to broaden the scope of retrieval capabilities. By integrating external knowledge sources like Wikipedia and Google Search APIs, RAG systems can access expansive real-time databases, enriching their generated content. For example, Toolformer~\cite{schick2024toolformer} taps into Wikipedia to access extensive knowledge bases, while the work by Lazaridou et al.~\cite{lazaridou2022internet} utilizes Google Search to ensure the most current and comprehensive information is incorporated, granting RAG systems a dynamic edge in content generation.

Overall, knowledge retrieval combines innovative methodologies with diverse data-sourcing strategies, underpinning the generation of informed and contextually relevant outputs.

\subsubsection{Knowledge Integration}
The knowledge integration phase in RAG frameworks is crucial for synthesizing coherent and accurate responses based on both retrieved and inherent knowledge. At this stage, researchers utilize LLM to generate the output and employ several preprocessing and efficiency strategies to improve its quality and efficiency.

The quality of retrieved content can drastically affect generation, with irrelevant or misleading data potentially having deleterious effects. LeanContext~\cite{arefeen2024leancontext} addresses this issue by utilizing reinforcement learning to selectively choose sentences that are most pertinent to the query, effectively minimizing the context size and reducing computational costs. Similarly, SELF-RAG~\cite{asai2023self} introduces a self-reflection mechanism where the language model assesses its own generated and retrieved content, allowing for corrective measures to be implemented during the generation process. SKR~\cite{wang2023self} proposes a dynamic framework whereby LLMs can rely on their pre-trained knowledge for recognized queries, resorting to retrieval only when necessary. FILCO~\cite{wang2023learning} enhances this by training a context-filtering model that screens out irrelevant data, thereby mitigating hallucination risks. Recent research by Lyu et al.~\cite{lyu2023improving} introduces an evaluation method specifically designed to assess the importance of the retrieved content. Their findings suggest that by pruning or re-weighting parts of the retrieval corpus, RAG systems can enhance their performance without requiring additional training rounds. 
In scenarios where the retrieval corpus is limited, approaches such as Selfmem~\cite{cheng2024lift} construct a memory pool from LLM-generated results, using an iterative selection framework to enhance generative quality. SAIL~\cite{luo2023sail} forms an instruction-tuning dataset based on retrieval outcomes, fine-tuning LLMs to ground responses in reliable content while excluding distracting elements.

Efficiency is another crucial concern, since encoding a large volume of retrieved passages is resource-intensive, leading to significant computation and memory overheads. To circumvent this, \cite{li2022decoupled} adopts a pre-encoding strategy that stores dense representations of the text corpus. LUMEN~\cite{de2023pre} takes it a step further by conditioning the encoder on current inputs and fine-tuning it specifically for this task to ensure high-quality pre-encoded memory representations. Innovative parallelism solutions, like PipeRAG~\cite{jiang2024piperag}, advance efficiency by enabling concurrent retrieval and generation, thus introducing alignment between retrieved content and real-time generation states. RAGCache~\cite{jin2024ragcache} further optimizes resource usage by caching intermediate states within a GPU-resident knowledge tree, employing a prefix-aware replacement policy to maintain the most critical key-value tensors. To efficiently compress retrieval contents, Xu et al.~\cite{xu2023recomp} introduce an extractive compressor to identify useful sentences and an abstract compressor to distill summaries.
Similarly, TCRA-LLM~\cite{liu2023tcra} utilizes a compression strategy, optimizing the semantic relevance of tokens to conserve computational resources.

In conclusion, the generation stage within RAG focuses on bolstering the quality and efficiency of LLM-driven content generation. These methodologies continually evolve to maximize the potential of LLMs in generating context-aware, accurate, and efficient responses for RAG.

\end{document}